\pdfoutput=1

\documentclass[11pt]{article}

\usepackage[]{EMNLP2023}

\usepackage{times}
\usepackage{latexsym}
\usepackage[edges]{forest}
\usetikzlibrary{arrows.meta}

\usepackage[T1]{fontenc}

\usepackage[utf8]{inputenc}
\usepackage{xcolor}
\definecolor{aqua}{HTML}{0096FF}
\definecolor{orange}{HTML}{FF9300}
\definecolor{salmon}{HTML}{FF7E79}
\definecolor{orchid}{HTML}{7A81FF}
\usepackage{amsmath}
\usepackage{amsfonts}
\usepackage{amssymb}
\usepackage{tikz}
\usetikzlibrary{trees, arrows.meta}

\usepackage{enumitem}

\usepackage{microtype}

\usepackage{inconsolata}

\usepackage{graphicx}

\usepackage{multirow}
\usepackage{booktabs}
\usepackage{xspace}

\newcommand{\taskname}{\textsc{GoalEx}\xspace}

\newcommand{\syndataset}{\textsc{Syn}\taskname{}\xspace}
\newcommand{\opendataset}{\textsc{Open}\taskname{}\xspace}
\newcommand{\methodname}{\textsc{PAS}\xspace}

\title{Goal-Driven Explainable Clustering via Language Descriptions}

\author{
    Zihan Wang\\
  \texttt{ziw224@ucsd.edu} \\\And
  Jingbo Shang\\
  \texttt{jshang@ucsd.edu} \\\And
  Ruiqi Zhong \\
  \texttt{ruiqi-zhong@berkeley.edu} \\}

\begin{document}
\maketitle
\begin{abstract}
Unsupervised clustering is widely used to explore large corpora, but existing formulations neither consider the users' goals nor explain clusters' meanings.
We propose a new task formulation, ``\textbf{Goal}-Driven Clustering with \textbf{Ex}planations'' (\taskname), which represents both the goal and the explanations as free-form language descriptions.
For example, to categorize the errors made by a summarization system, the input to \taskname{} is a corpus of annotator-written comments for system-generated summaries and a goal ``\textit{cluster the comments based on why the annotators think the summary is imperfect.}''; the outputs are text clusters each with an explanation (``\textit{this cluster mentions that the summary misses important context information.}''), which relates to the goal and accurately explains which comments should (not) belong to a cluster.
To tackle \taskname, we prompt a language model with ``[corpus subset] + [goal] + \textit{Brainstorm a list of explanations each representing a cluster.}'';
then we classify whether each sample belongs to a cluster based on its explanation; 
finally, we use integer linear programming to select a subset of candidate clusters to cover most samples while minimizing overlaps.
Under both automatic and human evaluation on corpora with or without labels, our method produces more accurate and goal-related explanations than prior methods.
\end{abstract}

\section{Introduction}
Text clustering is widely used to explore large corpora~\cite{aggarwal2012survey}.
However, existing formulations cannot adapt to different users' goals, which might be clustering based on sentiment, genre, or other properties~\citep{aharoni-goldberg-2020-unsupervised}; as a result, the desired output is under-specified. 
Furthermore, since the output clusters are not immediately interpretable, users must manually examine the clusters to gain insights. 
This can be time-consuming, especially when some clusters are semantically incoherent~\citep{chang2009reading}.

\begin{figure}[t!]
    \centering
    \includegraphics[width=\columnwidth]{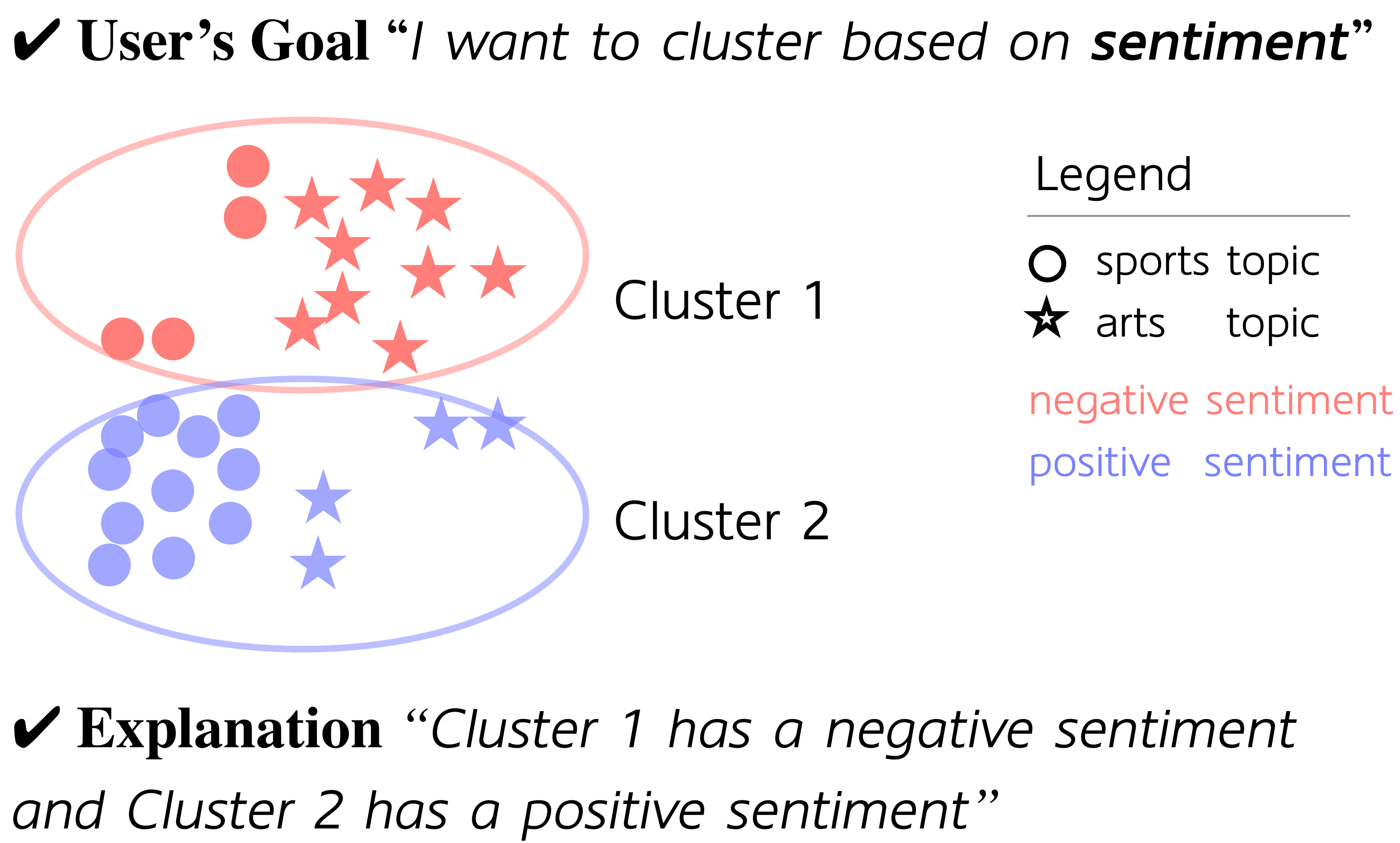}
    \caption{An illustration of our task formulation \taskname{} (\textbf{Goal}-Driven Clustering with \textbf{Ex}planations), where the input is a set of texts (corpus) and a goal, and the output constitutes a set of corpus subsets (clusters) each with an explanation. 
    Given the goal, a successful \taskname{} algorithm should cluster based on sentiment instead of topic and for each cluster explain which samples should (not) belong to it.
    }
    \label{fig:fig1}
    \vspace{-4mm}
\end{figure}

\begin{figure*}[t!]
    \centering
    \includegraphics[width=\textwidth]{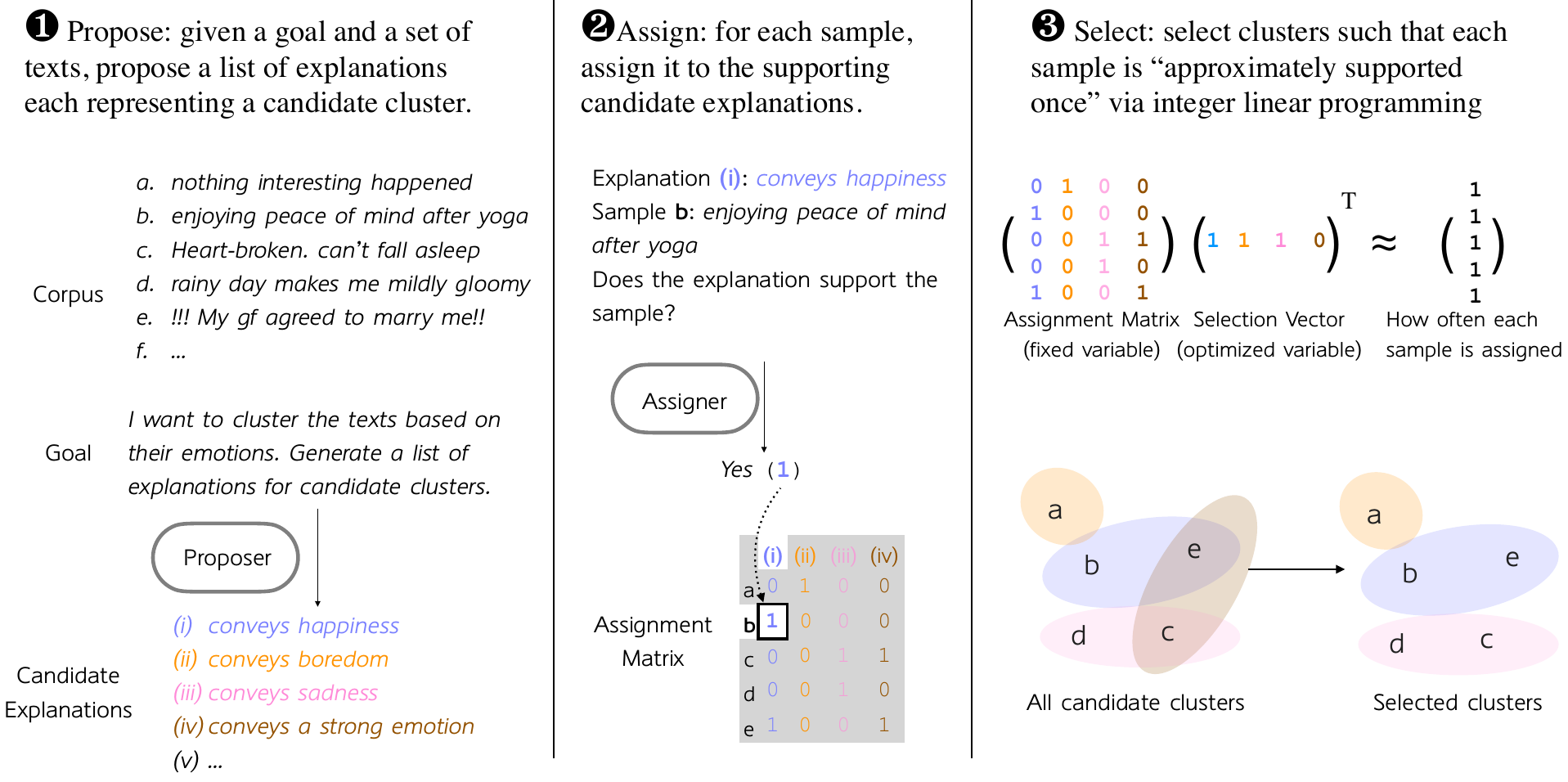}
    \caption{Our \textbf{P}ropose-\textbf{A}ssign-\textbf{S}elect (\methodname{}) algorithm to tackle the \taskname{} task. \textbf{Left, Propose}: We prompt a language model (``proposer'') with the goal and a subset of the corpus, obtaining a list of explanations for candidate clusters. \textbf{Middle, Assign}: we use a language model (``assigner'') to determine whether each explanation supports each sample.
    \textbf{Right, Select}: we use integer linear programming to select a subset of explanations, ensuring each sample has roughly one explanation,
    and obtain the selected set of clusters and explanations as our final output.}
    \label{fig:framework}
    \vspace{-6mm}
\end{figure*}

To address these weaknesses, we propose a new task formulation, \taskname{}, ``\textbf{Goal}-Driven Clustering with \textbf{Ex}planations'' (Section \ref{sec:task}). 
As illustrated in Figure \ref{fig:fig1}, the input to the task is a text corpus with multiple attributes (e.g., sports and arts related texts with different sentiments) and a goal description in natural language (``\textit{clustering based on sentiment}''). 
The output of the task constitutes a set of corpus subsets (clusters), each with a natural language explanation of which text samples should or should not belong to the cluster (e.g. ``\textit{contains positive sentiment}''). 
The output should satisfy three desiderata: 1) the explanations are goal-related, 2) each cluster is accurately described by its explanation, and 3) the clusters should overlap minimally while their union should cover most of the corpus.

To tackle \taskname{}, we develop a three-stage algorithm \textbf{P}ropose-\textbf{A}ssign-\textbf{S}elect (\methodname{}, Figure \ref{fig:framework}, Section \ref{sec:method}), each designed to address one of the desiderata.
At the proposal stage we address the 1st desideratum that the explanation should be goal-related;
we do this by prompting a language model (LM) to generate a list of goal-related explanations for candidate clusters based on the goal and a subset of the corpus.
At the assignment stage we address the 2nd desideratum that the explanations should accurately explain the clusters;
we do this by assigning text samples only to the explanations that support them.
At the selection stage we address the 3rd desideratum on maximizing coverage while minimizing overlap; we do this by using integer linear programming to search for a subset of candidate explanations so that each sample is roughly supported once.
At last, we output the selected explanations and their supported samples as clusters.

We benchmarked \methodname{} in two ways: 1) automatically evaluating its ability to recover known clusters from corpora (Section \ref{sec:benchmark}), and 2) manually evaluating its clusters and explanations on open-ended corpora (Section \ref{sec:open-ended}).
For automatic evaluation, we first compared \methodname{} to prior methods on recovering topic clusters underlying news and DBPedia articles and found that \methodname{} is competitive while additionally providing accurate explanations.
To test whether \methodname{} is goal-driven, we used an LM to synthesize a corpus, \syndataset, where each text has three known attributes: topic, language, and style;
\methodname{} effectively adapts to different goals such as ``\textit{clustering by topic / language / style}'', while prior methods fail catastrophically.

For open-ended evaluation, we constructed \opendataset{}, a collection of 12 open-ended \taskname{} problems from various NLP papers. 
We compared \methodname{} to previous clustering methods such as LDA and found that \methodname{}'s explanations are more accurate and goal-related under human evaluation.
Finally, we applied \methodname{} hierarchically to create progressively finer-grained clusters on \opendataset{}, inducing taxonomies over debate arguments, model errors, and customer reviews. \footnote{The implementation and data are made public in \url{https://github.com/ZihanWangKi/GoalEx}.}

Our contributions are summarized as follows. 
\begin{itemize}[leftmargin=*,nosep]
  \item We introduce \taskname{}, a novel setting for text clustering that takes into account of user's objectives and provides explanations for each cluster. 
  \item We developed the Propose-Assign-Select (\underline{\methodname{}}) algorithm and showed its effectiveness on established benchmarks.
  \item We tested \taskname{} to categorize debate points, customer feedback, and model inaccuracies in a hierarchical manner to show its potential to help users navigate extensive datasets effectively. 
\end{itemize}

\section{Defining \taskname{}} \label{sec:task}

We formalize the input-output space of \taskname{} and introduce the desiderata for an output.

\subsection{Input-Output Space}

\noindent The \textbf{input} of \taskname{} constitutes 

\begin{itemize}[leftmargin=*,nosep]
  \item a set of texts $X$ (the corpus);
  \item a string $g$ (the goal description);
  \item an integer $K$ (the desired number of clusters).
\end{itemize}

\noindent The \textbf{output} of \taskname{} constitutes 
\begin{itemize}[leftmargin=*,nosep]
  \item $K$ strings $e_{k}, k\in [K]$, where $e_{k}$ is an explanation of a cluster; additionally $e_{k}$ needs to be a natural language predicate that can be evaluated against an individual text sample;
  \item $K$ subsets of $X$: $C_{k} \subseteq X, k\in [K]$; each representing a cluster.
\end{itemize}

Note that goals and explanations can be arbitrary natural language strings and predicates much more complicated than the ones in Figure \ref{fig:fig1}. 
See examples in Section \ref{sec:open-ended}. 

\subsection{Desiderata}

We list three desiderata for a \taskname{} output, which inform our algorithm design in Section \ref{sec:method}. 

\noindent\textbf{Goal-Related.} The explanations should be goal-related.
For example, if a user's goal is to cluster based on sentiments, then ``\textit{has a positive sentiment}'' is goal-related, while ``\textit{is about sports}'' is not.

\noindent\textbf{Accurate Explanation.}
Since each explanation is a predicate, it should have a True evaluation on all samples from its corresponding cluster and False on others.
This automatically enforces the clusters to be semantically coherent, where the coherent interpretation is the explanation.

\noindent\textbf{Minimal Overlap and Maximal Coverage.} 
The clusters should overlap minimally while their union should cover most of the corpus samples. 
Ideally, every sample belongs to exactly one cluster.

\section{The Propose-Assign-Select Algorithm} \label{sec:method}

Each section between \ref{sec:propose} and \ref{sec:select} describes one stage of the \methodname{} algorithm (outlined in Figure \ref{fig:framework}).

\subsection{Propose Explanations for Clusters} \label{sec:propose}
The proposal stage aims to generate a list of $J$ (around 30-50) candidate explanations $\epsilon_{j}$.\footnote{We use $\epsilon_{j}$ to denote candidate explanations, while using $e_{k}$ to denote the final selected explanations}
We obtain them by prompting a language model (LM), which we refer to as the ``\textit{proposer}'' in the remaining text, to perform ``in-context clustering'' based on a random subset of the corpus;
concretely, the prompt concatenates $T$ samples from $X$, the goal $g$, and a request to generate $J'$ explanations for candidate clusters:

\vspace{1mm}
\noindent\textit{Sample 1. $x_{1};$ \dots \dots Sample $T$. $x_{T}$;}\\
\noindent\textit{Goal: g;}\\
\noindent\textit{Generate a list of $J'$ explanations for candidate clusters based on the samples.}
\vspace{1mm}

where we typically set the maximum $T$ such that the prompt length does not exceed the 75\% of the proposer's context window size and $J'=8 \ll T$.
The proposer would respond with a structured list of $J'$ candidate explanations:

\vspace{1mm}
\noindent \textit{Explanation 1. $\epsilon_{1};$\dots \dots Explanation $J'$. $\epsilon_{J'}.$}
\vspace{1mm}

Figure~\ref{fig:framework} left shows a more illustrative prompt-response pair.
Since the proposer's context window is usually not long enough to contain the entire corpus, we construct multiple prompts with different subsets from $X$ to allow the proposer to ``see'' as many different samples as possible during the proposal stage.
We sample from the proposer based on different prompts until obtaining $J$ explanations in total.
The full prompt is included in Appendix \ref{app:prompt-templates}.

\subsection{Assign Samples to the Correct Clusters} \label{sec:assign}
The assignment stage aims to determine whether each sample $x \in X$ is supported by each explanation $\epsilon_{j}$.
We determine this automatically by prompting an LM, which we refer to as the ``\textit{assigner}'':

\vspace{1mm}
\noindent``\textit{Predicate:} $\epsilon_{j}$\textit{. Text: }$x$. \textit{\\Is the Predicate true on the Text? Yes or No. When uncertain, output No.}''
\vspace{1mm}

We therefore obtain an \textit{assignment matrix} $\mathcal{A} \in \{0, 1\}^{|X| \times J}$, where $\mathcal{A}_{xj}$ indicates whether $x$ is supported by the $j^{\text{th}}$ explanation.\footnote{For convenience we also use $x$ as an index.}
Denote a candidate cluster as $C'_{\cdot} \subseteq X$, the $j^{\text{th}}$ candidate cluster is thus
\begin{equation}
    C'_{j} := \{x| x \in X, \mathcal{A}_{xj} = 1\}
\end{equation}

\subsection{Select an Optimal Subset of Clusters} \label{sec:select}
The selection stage aims to choose a subset of $K$ clusters from $J$ candidate clusters $C'_{j}$, so that each sample $x$ belongs to roughly one selected cluster.

Define the selection vector $s \in \{0, 1\}^{J}$ to be a row vector, where $s_{j}$ indicates whether $C'_{j}$ is selected.
Since we require $K$ selected clusters, we add the constraint:
\begin{equation} \label{eq:cluster-count}
    s \cdot \mathbf{1} = K 
\end{equation}

We introduce a row vector variable $m$
\begin{equation} \label{eq:inclusion-count}
    m := \mathcal{A}s^{T} \in \mathbb{N}^{|X|},
\end{equation}
where $m_{x}$ counts how many selected clusters include $x$.
An ideal $s$ should result in $m_{x} = 1$ for all $x$, since $m_{x} > 1$ implies that at least two selected clusters overlap on $x$ while $m_{x} < 1$ implies $x$ is ``missed'' by all clusters. 
Therefore, we design the following loss function $f_{\lambda}$ to track how much an entry from $m$ diverges from 1:
\begin{equation}
f_{\lambda}(m_{x}) :=
\begin{cases}
    (1-m_{x}) & \text{if } m_{x} < 1 \text{, ``miss'';} \\
    0 & \text{if } m_{x} = 1 \text{, ``ideal'';}\\
    \lambda(m_{x}-1) & \text{if } m_{x} > 1 \text{, ``overlap'';}
\end{cases}
\end{equation}
where $\lambda$ is a hyper-parameter determining how much overlaps are penalized.
To conclude, we will minimize the following loss $\mathcal{L}$ for $s$
\begin{equation} \label{eq:non-linear-loss}
    \mathcal{L}(s) := f_{\lambda}(m) \cdot \mathbf{1},
\end{equation}
subject to the constraint of Equation \ref{eq:cluster-count} and \ref{eq:inclusion-count}. 

However, it is hard to directly minimize this loss as written because it requires searching over discrete variables under a piecewise-linear loss. 
Therefore, we reduce it to an integer linear programming (ILP) problem, which can be effectively solved\footnote{efficiently find a reasonable solution empirically even though it is theoretically NP-Complete} by existing libraries.
To perform the reduction, we introduce an auxiliary row vector variable $a \in \mathbb{R}^{|X|}$ and add the following two constraints
\begin{equation}\label{eq:auxillary}
    a \succcurlyeq 1 - m, a \succcurlyeq \lambda (m - 1),
\end{equation}
where $\succcurlyeq$ denotes element-wise greater or equal to. 
To conclude, we will minimize the loss $\mathcal{L}$
\begin{equation} \label{eq:final-loss}
    \mathcal{L} = a\cdot \mathbf{1},
\end{equation}
subject to the constraints in Equation \ref{eq:cluster-count}, \ref{eq:inclusion-count}, and \ref{eq:auxillary}, which are all linear. 
We explain our implementation in python code with comments in Appendix \ref{app:selection-pseudo}.
We refer to one sequential application of propose, assign, and select as one \textit{iteration} of \methodname{}. 

In addition to the three stages above, \methodname{} involves other auxillary procedures such as 1) running \methodname{} for 5 iterations to cover the entire corpus, and 2) committing each sample to one single cluster when needed.
Due to space constraints, we outline other auxiliary steps of \methodname{} in Appendix \ref{sec:other}.



\section{Automatic Evaluation} \label{sec:benchmark}
Following the evaluation protocol from prior works, we evaluated \methodname{} by applying it to corpora that are mixtures of known clusters, treating the known clusters and their explanations as the reference solutions, and checking how well the outputs of \methodname{} can recover the references.
We evaluated \methodname{} under two settings: traditional topic clustering and goal-driven non-topic clustering.
In both settings, we compared 1) the similarity between the reference and the output clusters automatically and 2) the similarity between the explanations manually.
We found that \methodname{} is comparable to previous methods for topic clustering and recovers most of the reference explanations; additionally, since \methodname{} is goal-driven, it performs significantly better when there are multiple ways to cluster a corpus.

The following sections will present the datasets (Section \ref{sec:dataset}), the clustering methods we evaluated (Section \ref{sec:method-evaluated}), the evaluation protocol (Section \ref{sec:metrics}), and the performance of each method (Section \ref{sec:performance}). In addition, we evaluated the quality of each stage of \methodname{} in Appendix~\ref{app:per-stage}.

\subsection{Datasets} \label{sec:dataset}
We evaluated on both corpora from prior works for topic clustering and other corpora for non-topic clustering.
We considered four datasets: (AG)'s News, (DB)pedia, (NYT) News, and (SYN)\taskname{}.
We use ($\cdot$) to denote a dataset abbreviation in this section.

\noindent\textbf{(AG)}'s News~\cite{zhang2015character} is a news topic classification dataset with four topic clusters: politics, sports, business, and technology.

\noindent\textbf{(DB)}pedia~\cite{zhang2015character} is a corpus of  articles classified into ontologies, such as athlete and book, with 14 topic clusters in total.

\noindent\textbf{(NYT)} News~\cite{meng2020discriminative} is a corpus of New York Times news articles, each with a topic label and a location label. 
There are in total 9 topics, e.g., politics, arts, and 10 locations, e.g., France, Italy. 
We subsampled this corpus so that the topic and location labels are balanced.

\noindent \textbf{(SYN)}\taskname{} To test \methodname{}'s ability to cluster based on different goals, we synthesized (SYN)\taskname{}, which can be clustered based on three different dimensions: Topics, Writing Style, or Language.
To synthesize \syndataset{}, we first designed four values for each dimension, e.g. ``1.\textit{French}''/2.``\textit{English}''/3.``\textit{Spanish}''/4.``\textit{Deutsch}'' for the Language dimension.
Then we took the Cartesian product across three dimensions, obtaining 4$^3$=64 value combinations; for example, one combination could be ``\textit{Language: French, Style: Poem, Topic: Sports}''.
Finally, for each of the 64 value combinations, we prompted \texttt{Claude-v1.3} to generate 16 text samples conditionally on the values, resulting in 1024 samples for \syndataset{} in total.
Therefore, the reference clusters are different if we cluster based on different dimensions, hence penalizing methods that ignore the goals.
Appendix \ref{app:syngoal} includes more details about the values for each dimension and the prompt we used for conditional generation.

The first three datasets might have appeared in the pre-training corpus of \texttt{gpt-3.5-turbo}, thus raising potential concerns about memorization. We believe our task of proposing explanations on the three datasets did not occur in the pre-training corpus, thus justifying the validity of our evaluations. A more detailed justification is in Appendix \ref{app:memorization}.

\subsection{Methods and Baselines} \label{sec:method-evaluated}
We compared fours methods: \underline{\methodname{}}, \underline{LDA}, \underline{E5}, and \underline{Instructor}.
For all methods, we set the number of clusters to be that of the reference solution.
We use \underline{\phantom{0}$\cdot$\phantom{0}} to denote a method in this section.

\noindent\textbf{\underline{\methodname}} is described in Section \ref{sec:method}. 
By default, we used \texttt{gpt-3.5-turbo} as the proposer and \texttt{flan-t5-xl} ~\citep{chung2022scaling} as the assigner. 
We set $J = 30$ and $\lambda = 0.5$, except for the (DB)pedia dataset and (NYT) News dataset where we set $\lambda = 0.3$ since they have many target clusters.
We additionally require each $x$ to appear in exactly one cluster using the commitment method described in Section \ref{sec:other}. 

\noindent\textbf{\underline{LDA}}~\cite{blei2003latent}, or Latent Dirichlet Allocation, is a standard generative probabilistic model that identifies hidden topic clusters in a corpus by assuming that each text is a mixture of topics and each topic is a distribution of words.

\noindent\textbf{\underline{Instructor}}~\cite{su2022one} is contrastively trained on a large collection of datasets with annotated instructions;
as a result, it can create specialized text embeddings according to the instructions.
To perform goal-driven clustering, we rephrased our goal as the embedding instruction.
We computed the text embeddings with \texttt{instructor-xl} and then ran K-means clustering.

\noindent\textbf{\underline{E5}}~\cite{wang2022text} is a contrastively trained text embedder on crawled data, e.g. post-comment pairs and annotated data, e.g. NLI. We computed the text embeddings with \texttt{e5-large} and then ran K-means to obtain the clusters.

Appendix~\ref{app:implementation} includes more implementation details, e.g. what library we used.

\subsection{Metrics} \label{sec:metrics}
We follow the standard protocol from ~\citet{lange2004stability} to evaluate a clustering method, where we first match each of the output cluster to a known reference cluster and then compute the similarity of each pair of matched clusters via F$_1$ score.
Denoting the $k^{\text{th}}$ output cluster as $\hat{C}_{k'}$ and the $k^{\text{th}}$ reference as $C^{*}_{k}$,
we formulate the matching problem as a bipartite matching problem and solve it with Hungarian algorithm, where the edge weight between each pair of reference and output cluster is the size of their overlap, $|\hat{C}_{k'} \cap C^{*}_{k}|$.
After matching finishes, for each pair of matched reference and output clusters $C^{*}$ and $\hat{C}$, we compute the F$_1$ score of predicting whether $x \in C^{*}$ based on whether $x \in \hat{C}$, and then average across all $k$ to compute the final macro F$_1$ score for evaluation.

\subsection{Results} \label{sec:performance}

We evaluated \underline{\methodname{}} on topic clustering and goal-driven clustering based on other dimensions. All results shown below are the average of 3 trials with different random seeds.

\noindent \textbf{Recovering Topic Clusters.} We first evaluated the clustering methods on recovering topic clusters and report the results in Table~\ref{tab:topic_clustering}. 
\underline{\methodname} consistently outperforms \underline{LDA}.
\underline{\methodname} slightly outperforms  \underline{Instructor} on (AG), (NYT), and (SYN);
on (DB), \underline{\methodname} is underperforming \underline{Instructor} by 11\%.

\begin{table}[t]
    \small
    \centering
    \begin{tabular}{l|cccc}
    \toprule
    \textbf{} Macro F$_1$ (\%) & (AG) & (DB) & (NYT) & (SYN) \\
    \hline
    \underline{Random}        & 27 & 11 & 14 & 27 \\
    \underline{LDA}           & 53 & 51 & 51 & 28 \\
    \underline{E5}            & 86 & 72 & 67 & 96 \\
    \underline{Instructor}    & 84 & \textbf{82} & 69 & 77 \\
    \hline
    \underline{\methodname}   &\textbf{ 87} & 71 & \textbf{70} & \textbf{98} \\
    \bottomrule
    \end{tabular}
    \caption{We compare different methods and \underline{\methodname} for recovering topic clusters and report the macro F$_1$ score for each method, along with a random baseline which assigns each sample to a cluster randomly.
    }
    \label{tab:topic_clustering}
    \vspace{-4mm}
\end{table}
\begin{table}[]
    \centering
\begin{tabular}{llr}
\toprule
label  & output &  F$_1$ \\
\midrule
                ``\textit{\phantom{} Company}'' &                ``\textit{\phantom{} business}'' &  75 \\
               ``\textit{\phantom{} Building}'' &            ``\textit{\phantom{} architecture}'' &  82 \\
                 ``\textit{\phantom{} {\color{red}Animal}}'' &       ``\textit{\phantom{} {\color{red}lakes}}'' &   3 \\
                  ``\textit{\phantom{} {\color{red}Plant}}'' &                 ``\textit{\phantom{} {\color{red} biology}}'' &  73 \\
           \dots & \dots & \dots \\
\bottomrule
\end{tabular}
    \caption{We ran \underline{\methodname{}} on (DB)pedia to cluster based on topics and present its cluster explanations. 
    We abbreviate each explanation by removing the prefix ``\textit{has a topic of}'' (e.g., ``\textit{\phantom{} Company}'' corresponds to a full explanation ``\textit{\phantom{} has a topic of Company.}''). 
    The table of all 14 explanations is in Appendix Table~\ref{tab:app-descriptions-dbpedia}.}
    \label{tab:descriptions-dbpedia}
    \vspace{-2mm}
\end{table}

To understand why our method does not deliver the best performance on (DB), we manually examined the explanation (for one of the three trials) for each output cluster and present it in Table \ref{tab:descriptions-dbpedia} along with its matching reference.
Overall, the outputs are similar to the references;
the performance drop is mainly because \underline{\methodname{}} completely missed the ``\textit{Animal}'' cluster, since it ``merged'' it with the ``\textit{Plant}'' reference cluster into a ``\textit{Biology}'' cluster.
Additional evidence on this merging effect can be found in Appendix~\ref{app:per-stage}.
We consider such a mistake benign and hence conclude that \underline{\methodname{}} is on par with previous state of the art on topic clustering. 

Next, we will show that \underline{Instructor} fails catastrophically on non-topic based clustering, implying that it has an ``inductive bias'' to cluster on topics.


\begin{table}[t]
    \small
    \centering
    \begin{tabular}{l|ccc}
    \toprule
    \multirow{2}{*}{Macro F$_1$ (\%)} & \multicolumn{1}{c}{(NYT)} & \multicolumn{2}{c}{(SYN)} \\
     & Location  & Language & Style \\
    \hline
    \underline{Random}        & 13  & 27 & 27 \\
    \underline{LDA}           & 40  & 83 & 25 \\
    \underline{E5}            & 55  & 27 & 27 \\
    \underline{Instructor}    & 54  & 25 & 25 \\
    \hline
    \underline{\methodname}   & \textbf{76}  & \textbf{97} & 31  \\
    \underline{\methodname}$\dagger$    & - &  - & \textbf{45}  \\
    \bottomrule
    \end{tabular}
    \caption{We ran \methodname{} to cluster based on non-topic goals and report the macro F$_1$ score. $\dagger$ We use \texttt{gpt-4} as the proposer and \texttt{gpt-3.5-turbo} as the assigner.
    }
    \label{tab:nontopic_clustering}
    \vspace{-6mm}
\end{table}

\noindent \textbf{Recovering Non-Topic Clusters.}
We now evaluate \underline{\methodname{}} on other goals and report the performance in Table \ref{tab:nontopic_clustering} --- specifically, clustering based on locations on (NYT) and writing styles or languages on (SYN).
Since \underline{\methodname{}} is goal-driven, it performs significantly better than previous methods.

\begin{table*}[]
    \centering
\begin{tabular}{lll}
\toprule
                             reference  &                                        output  &  F$_1$ \\
\midrule
    ``\textit{has a writing style of twitter}'' & ``\textit{has a writing style of instructional or informat...}'' &  45 \\
``\textit{has a writing style of screen play}'' & ``\textit{has a writing style of narrative or storytelling...}'' &  51 \\
        ``\textit{has a writing style of rap}'' & ``\textit{has a writing style of using rhymes and rhythm}'' &  49 \\
       ``\textit{has a writing style of poem}'' & ``\textit{has a writing style of incorporating foreign lan...}'' &  34 \\
\bottomrule
\end{tabular}
    \caption{We ran \underline{\methodname{}} with proposer=\texttt{gpt-4} and assigner=\texttt{gpt-3.5-turbo} to cluster (SYN)\taskname{} based on Style. 
    We present the four output explanations and compare them to the references. 
    Although our method is still far from perfect, the \underline{\methodname{}} is able to generate similar explanations for the 2nd and 3rd row.
    }
    \label{tab:tbl-descriptions-syn-style-gpt4}
    \vspace{-4mm}
\end{table*}

However, \underline{\methodname{}} with our default configuration is poor at writing style clustering on (SYN).
Fortunately, the performance can improve significantly by 14\% after using more capable models -- \texttt{gpt-4} \citep{openai2023gpt4} as the proposer and \texttt{gpt-3.5-turbo} as the assigner.
We present the output explanations in Table~\ref{tab:tbl-descriptions-syn-style-gpt4} and expect \underline{\methodname{}} to improve with future better LMs.





\noindent \textbf{Sensitivity Study.} 
We conducted a prompt sensitivity study in Appendix~\ref{app:sensitivity} and concluded that our method is not sensitive to the prompts we chose for the proposer and assigner. 
We also conducted a dataset sensitivity study on (DB)pedia to study how imbalance of classes in the dataset or noisy, out-of-distribution data points would affect our algorithm, and concluded that our algorithm is not especially vulnerable to these noises than \underline{Instructor}.

\noindent \textbf{Ablation Studies for \underline{\methodname{}}. } 
Finally, we conducted two ablations for \underline{\methodname{}} to study the contribution of 1) proposing multiple iterations and 2) our selection algorithm.
We present the results in Appendix \ref{app:further-ablation} and found that running \underline{\methodname{}} for five iterations improves over one iteration and using an ILP algorithm with a positive $\lambda$ improves the performance.

\section{Open-Ended Advanced Applications} \label{sec:open-ended}
While \underline{\methodname{}} achieves high performance on benchmarks with cluster labels, it does not necessary imply high performance under real applications.
Therefore, we constructed \opendataset{}, a collections of 12 open-ended realistic \taskname{} problems to evaluate \underline{\methodname{}}. 
Since these problems do not have cluster labels, we evaluated \underline{\methodname{}} with the three metrics introduced in Section \ref{sec:task}: (1) explanation accuracy, (2) goal-relevance, and (3) coverage and overlap.
As (1) and (2) require human annotators and are hence expensive to conduct repeatedly, we used them to test the limit of \underline{\methodname{}} to inform future research:
we applied \underline{\methodname{}} with the highest quality models under our budgetary constraints and challenged it to generate taxonomies by producing trees of progressively finer-grained clusters on \opendataset{}.
We evaluated \underline{\methodname{}} quantitatively with human annotators for the first layer of the taxonomy and qualitatively analyzed the rest. 

\subsection{\opendataset{}}
To evaluate under real applications, we constructed \opendataset{}, a collection of 12 open-ended \taskname{} problems.
Each corpus comes from an NLP paper or a Kaggle website and we annotated it with a goal related to the paper. 
For example:

\begin{itemize}[leftmargin=*,nosep]
  \item  comments for model-generated summaries \cite{scheurer2023training}, with the goal of ``\textit{categorizing model errors}
  \item debates on why spanking is bad, with the goal of ``\textit{categorizing the types of arguments}'' \cite{habernal-gurevych-2016-makes}
\end{itemize}

Appendix \ref{sec:opengoalex} includes all 12 problem descriptions and citations.
To reduce reporting bias, we collected \opendataset{} before our experiments.

\subsection{Advanced Application of \underline{\methodname{}}}

To generate a taxonomy for each corpus, we first apply \underline{\methodname{}} for the entire corpus;
then for every output cluster with  $>20$ samples, we apply \underline{\methodname{}} again to create finer-grained clusters and output trees of explanations as taxonomies;
when creating child clusters for a parent cluster, we include the explanation for the parent into the original goal and request the new candidates to be sub-categories of the parent's explanation. 
Here is an example goal where the parent's explanation is in \textbf{bold}:

\noindent ``\textit{My goal is to cluster comments for model-generated summaries falling under the following category: \textbf{whether this comment advises adding omitted details; specifically, \dots. For example, \dots'}, and I want to create finer-grained cluster descriptions that fall under the above category.}''

We set $K = 8$, $\lambda = 0.5$, proposer $=$ \texttt{gpt-4}, and assigner $=$ \texttt{Claude-v1.3} \citep{bai2022constitutional}. 
We allow a sample to appear in multiple clusters so that \underline{\methodname{}} can see as many samples as possible when creating subcategories.
We designed a new prompt template to propose more detailed explanations; see Appendix Figure \ref{fig:proposal-complex} for more details.

\subsection{Quantitative Evaluation} \label{sec:open-ended-eval}
We quantitatively evaluated the first layer of taxonomy (i.e. the output of the standard \taskname{} formulation) based on the three metrics introduced in Section \ref{sec:task}.
To help the readers interpret our results, we compared \underline{\methodname{}} to \underline{LDA} and \underline{Instructor}.

\begin{figure}
    \centering
    \includegraphics[width=\columnwidth]{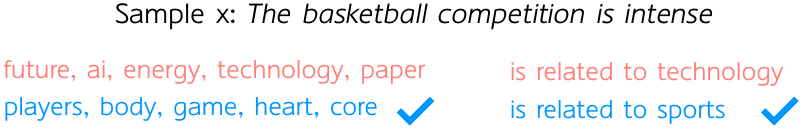}
    \caption{Explainability evaluation instances. Given the sample $x$, the evaluator needs to choose which one of the explanations (blue, model generated explanation for $x$, or orange, a random different model generated explanation) is more related. \textbf{Left}: top-word based explanations by \underline{LDA} and \underline{Instructor}. \textbf{Right}: natural language explanations by \underline{\methodname{}}.}
    \label{fig:expl}
    \vspace{-4mm}
\end{figure}

\paragraph{Explanation Accuracy.}
If an explanation $e_{k}$ is accurate for cluster $C_{k}$, then given a sample $x \in C_{k}$ in its cluster, a human should be able to tell whether $e_{k}$ or $e_{k'}$, the explanation for another cluster $C_{k'}$, is more related.
We call the tuple $(x, e_{k}, e_{k}')$ an explainability evaluation instance and show an example in Figure \ref{fig:expl}. 
To sample an instance, we randomly sampled a problem from \opendataset{}, sample an output cluster $C_{k}$, and then sampled a text sample $x \in C_{k}$; we then randomly sample a distractor explanation $e_{k'}$ such that $x \notin C_{k'}$. 
For each instance, we present it to three human turkers and consider it correct if the majority of them choose $e_{k}$ over $e_{k'}$. 
We include more details for this HIT task and how to generate word-based explanations for \underline{LDA} and \underline{Instructor} in Appendix \ref{app:expl-instance}.


We ran study with Turkers and found that they can choose the corresponding explanation 80\% of the time for \underline{\methodname{}}, outperforming 56\% for LDA ($p \approx 10^{-9}$) and 71\% for \underline{Instructor} ($p < 10^{-3}$).\footnote{Our evaluation weighted each cluster explanation uniformly, so these results imply that ``\underline{\methodname{}} produces on average more accurate explanations'', but not ``each sample $x$ is more accurately explained''.}

\noindent \textbf{Relevance.}
We evaluated how well \underline{\methodname{}}'s explanations relate to the goal and compared them to explanations for \underline{LDA} and \underline{Instructor} clusters.
For each problem in \opendataset{}, we randomly sampled a problem, an explanation from \underline{\methodname{}}'s output, and one from a baseline approach;
we then asked the evaluators to choose which explanation is more relevant to the goal, or abstain if they are similar.
To ensure reliability and fairness of our evaluation, the authors performed evaluations on their own rather than relying on Turkers, since the goals in \opendataset{} are technical and motivated by NLP research; the evaluators are also unaware of whether the baselines or \underline{\methodname{}} generated each explanation: to make the baseline explanations stylistically similar to \underline{\methodname{}}'s outputs, we used the \textsc{D5} system by \citet{zhong2023goal} to describe the differences between each cluster and the rest of the corpus in natural language.


Table \ref{tab:relevance} reports the results for a direct pair-wise comparisons between \underline{\methodname{}} and \underline{LDA}/\underline{Instructor}. 
\underline{\methodname{}}'s explanations are more often related to the goal compared to \underline{LDA} ($p$-value < $10^{-3}$) and \underline{Instructor} ($p$-value < 0.05), which are not goal-driven. 
As a robustness check, two authors independently reproduced the exact same conclusion.

\begin{table}[]
    \centering
    \begin{tabular}{lccc}
        \toprule
        Baseline & Win (\%) & Lose (\%) & $p$-value \\
        \hline
        \underline{Instructor} &  30 & 12 & < $10^{-3}$ \\
        \underline{LDA} & 51 & 13 & < $10^{-8}$ \\
        \hline
    \end{tabular}
    \caption{How often is \underline{\methodname{}}'s explanations are more relevant compared to the baselines.}
    \label{tab:relevance}
    \vspace{-4mm}
\end{table}

\noindent \textbf{Coverage and Overlap.}
On average, 66\% of the text samples are covered by at least one cluster, and 60\% of the samples are covered by one and only one cluster;
this is one of the key limitations of the current \underline{\methodname{}} system, as traditional clustering methods such as \underline{LDA} or \underline{Instructor} would cover 100\% samples exactly once.
On the flip side, however, these low numbers might reflect the inherent difficulty of producing semantically coherent clusters with $K$ = 8 in a goal-driven way;
when a sample cannot be supported by any explanation, it might actually be better to explicitly consider it ``not covered'' as our approach does, rather than forcing it to a semantically incoherent cluster and creating a delusion of 100\% coverage.

\begin{figure*}[t!]
    \centering
    \includegraphics[width=\linewidth]{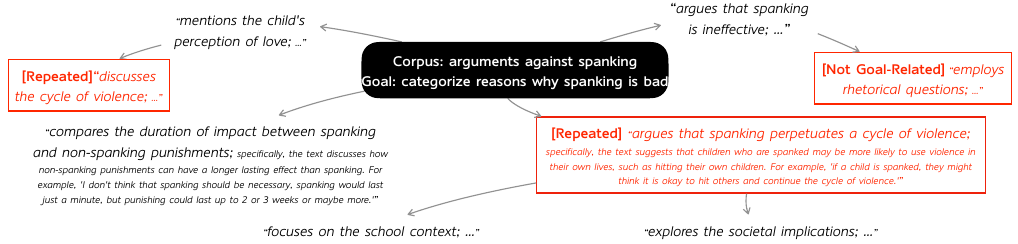}
    \caption{Example taxonomy for arguments against spanking \citep{habernal-gurevych-2016-makes} produced by \methodname{}. 
    }
    \label{fig:spanking}
    \vspace{-4mm}
\end{figure*}

\subsection{Qualitative Analysis} \label{sec:qualitative}

We show one subpart of an example taxonomy of `\textit{`why spanking is bad}'' in Figure \ref{fig:spanking} to obtain a qualitative understanding of \methodname{}.
Most explanations are goal-related and they could help the users quickly explore the corpus without inspecting each cluster manually;
however, some do not form a coherent taxonomy. 
For example, the explanation ``\textit{employs rhetorical questions}'' is irrelevant to the goal of identifying argument types;
additionally, the explanation ``\textit{discusses the cycle of violence}'' appears both in the first and second levels of the taxonomy and hence should be merged.
We present example taxonomies over customer complaints and model errors in Appendix Figure \ref{fig:demo} and \ref{fig:product}.

\section{Related Work}

\noindent\textbf{Text Clustering.}
Most existing text clustering methods first encode each text sample into some vector and then run a clustering algorithm; e.g. one hot bag-of-words encodings and tf-idf~\cite{blei2003latent, aggarwal2012survey}, or neural word/context embeddings (\citet{aharoni2020unsupervised,wang2022text,su2022one}, \textit{inter alia}).
Using text clustering methods as backbones, many prior works such as \citet{luu-etal-2014-taxonomy, shang2020nettaxo, downey-etal-2015-efficient} apply them hierarchically to a corpus to produce taxonomies over topics.
Nevertheless, previous text clustering algorithms do not necessarily produce interpretable clusters \citep{chang2009reading}, mostly studies topic clustering, and cannot flexibly adapt to users' goal.

\noindent\textbf{Explaining Text Clusters.}
To explain topic clusters, \citet{carmel2009enhancing} proposes to explain each cluster by extracting candidate labels either from text or from Wikipedia;
\citet{treeratpituk2006automatically} proposes to explain each cluster by selecting candidate labels using information from the cluster, the parent cluster, and corpus statistics;
\citet{zhang2018taxogen} proposes to summarize a cluster with a group of semantically coherent phrases. 
However, these solutions are limited, since phrase-level explanations are not flexible enough to describe a complex cluster.
\citet{zhong2022describing} proposes to explain a text cluster by describing its differences with the rest of the corpus in natural language; however, its explanation usually does not fully cover the entire cluster, while our clusters are explainable by construction during the assignment stage.

\noindent\textbf{Controlling the Clustering Process.}
We need additional supervision signals from the users so that they can have more control over the clustering process. 
\citet{hu2014interactive} allows the users to shape the clusters by specifying words that should co-occur in a cluster.
In the image domain, Open World Classification (OWC)~\citep{shu2018unseen,cao2021open}, also known as Generalized Category Discovery~\cite{vaze2022generalized}, gives the users more control by asking for a few example labels and their example datapoints;
for example, given five labels and some corresponding images in a CIFAR10 dataset ~\citep{krizhevsky2009learning} (e.g., ``automobile'', ``bird'', etc), discover the remaining five labels on the unlabeld dataset (e.g., ``frog'', ``ship'') and classify the entire dataset into 10 labels ~\citep{zheng2022towards,zhao2023incremental,xie2023open}; closest to our work, \citet{wang2023wot} operates OWC in the text domain.
Our work proposes a complementary direction and allows the user to control the clustering process with a goal description, which is more expressive and lightweight.



\noindent\textbf{Explaining Patterns via Language.}
Natural language can be used to help users explain patterns in text data ~\citep{zhong2022describing, singh2022explaining}.
With the increasing capability of language models \cite{openai2023gpt4}, they are used to explain more complicated patterns, such as the inner workings of neural networks ~\citep{singh2023explaining, Bills_Cammarata_Mossing_Tillman_Gao_Goh_Sutskever_Leike_Wu_Saunders_2023}.
Our system is closest to D5 developed by ~\citet{zhong2023goal}, which describes difference between text distributions in a goal-driven way. 

Patterns in other modalities can also be described via language.
For example, \citet{zhu2022gsclip} describes distribution shifts for image classification tasks and \citet{eyuboglu2022domino} describes errors made by vision models.
With future advances in multi-modal foundation models, we hope that \taskname{} can be extended to cluster images, and potentially sound \citep{aghajanyan2023scaling} or physical senses \citep{thomason2016learning}.


\section{Conclusion}
We proposed \taskname{}, a new formulation for text clustering that adapts to a user's goal and outputs explanations for each cluster.
To tackle \taskname{}, we developed the Propose-Assign-Select (\underline{\methodname{}}) algorithm; 
under automatic evaluation with known references and open-ended applications, \methodname{} can generate accurate and goal-related explanations for the clusters.
Finally, we applied \taskname{} hierarchically to produce taxonomies over debate arguments, customer complaints, and model errors, thus assisting users to explore large corpora. 
Future works can improve on discovering minority clusters, following the goal better, and resolving global inconsistency when applying \methodname{} recursively.


\section*{Limitation}

As indicated in Section \ref{sec:qualitative}, \methodname{} cannot yet construct coherent taxonomies.
As indicated in Section \ref{sec:open-ended-eval}, \methodname{} is far from being able to cover all the samples and the clusters have significant overlap. 
Given these weaknesses, a practitioner should still properly interpret the results of PAS.

Our evaluation is also not universal in scope. 
Our benchmarks are predominantly in English, and hence our results do no necessarily generalize to other languages.
Our dataset \opendataset{} also implicitly encodes the author's biases for what clustering problems are more important than the other, though this is a universal problem for any newly proposed benchmark.
We hope that with a combination of automatic evaluation on datasets from prior work and human evaluation on open-ended \taskname{} problems that we collected, we can more robustly, though not perfectly, establish the conclusions from our paper.
We also did not evaluate our methods under situations where the number of clusters $K$ is large, e.g., $K$ > 50.

Finally, reaching the best performance requires using \texttt{gpt-4} and \texttt{claude-v1.3} as the proposer and the assigner, which might induce a large cost via LM-APIs if one needs to run PAS on a large corpus; we hope such a problem would alleviate in the future if we could use a lighter weight model to approximate the assigner, the cost of computation significantly decreases, or there is a more computationally efficient variant of PAS.

\section*{Ethics Statement}

The human evaluation is approved by the Institutional Review Board. 

\section*{Acknowledgement}
Our work is sponsored in part by NSF CAREER Award 2239440, NSF Proto-OKN Award 2333790, NIH Bridge2AI Center Program under award 1U54HG012510-01, Cisco-UCSD Sponsored Research Project, as well as generous gifts from Google, Adobe, and Teradata. Any opinions, findings, and conclusions or recommendations expressed herein are those of the authors and should not be interpreted as necessarily representing the views, either expressed or implied, of the U.S. Government. The U.S. Government is authorized to reproduce and distribute reprints for government purposes not withstanding any copyright annotation hereon. Ruiqi Zhong is funded by NSF-Simons Theorinet Grant (NSF Award \#2031985). We thank members from Jacob Steinhardt's group, Jingbo Shang's group, and Berkeley NLP group for paper feedback.

\bibliography{anthology,custom}

\begin{thebibliography}{53}
\expandafter\ifx\csname natexlab\endcsname\relax\def\natexlab#1{#1}\fi

\bibitem[{Aggarwal and Zhai(2012)}]{aggarwal2012survey}
Charu~C Aggarwal and ChengXiang Zhai. 2012.
\newblock A survey of text clustering algorithms.
\newblock \emph{Mining text data}, pages 77--128.

\bibitem[{Aghajanyan et~al.(2023)Aghajanyan, Yu, Conneau, Hsu, Hambardzumyan,
  Zhang, Roller, Goyal, Levy, and Zettlemoyer}]{aghajanyan2023scaling}
Armen Aghajanyan, Lili Yu, Alexis Conneau, Wei-Ning Hsu, Karen Hambardzumyan,
  Susan Zhang, Stephen Roller, Naman Goyal, Omer Levy, and Luke Zettlemoyer.
  2023.
\newblock Scaling laws for generative mixed-modal language models.
\newblock \emph{arXiv preprint arXiv:2301.03728}.

\bibitem[{Aharoni and
  Goldberg(2020{\natexlab{a}})}]{aharoni-goldberg-2020-unsupervised}
Roee Aharoni and Yoav Goldberg. 2020{\natexlab{a}}.
\newblock \href {https://doi.org/10.18653/v1/2020.acl-main.692} {Unsupervised
  domain clusters in pretrained language models}.
\newblock In \emph{Proceedings of the 58th Annual Meeting of the Association
  for Computational Linguistics}, pages 7747--7763, Online. Association for
  Computational Linguistics.

\bibitem[{Aharoni and Goldberg(2020{\natexlab{b}})}]{aharoni2020unsupervised}
Roee Aharoni and Yoav Goldberg. 2020{\natexlab{b}}.
\newblock Unsupervised domain clusters in pretrained language models.
\newblock \emph{arXiv preprint arXiv:2004.02105}.

\bibitem[{Asai et~al.(2018)Asai, Evensen, Golshan, Halevy, Li, Lopatenko,
  Stepanov, Suhara, Tan, and Xu}]{happy-moments}
Akari Asai, Sara Evensen, Behzad Golshan, Alon Halevy, Vivian Li, Andrei
  Lopatenko, Daniela Stepanov, Yoshihiko Suhara, Wang-Chiew Tan, and Yinzhan
  Xu. 2018.
\newblock Happydb: A corpus of 100,000 crowdsourced happy moments.
\newblock \emph{arXiv preprint arXiv:1801.07746}.

\bibitem[{Bai et~al.(2022)Bai, Kadavath, Kundu, Askell, Kernion, Jones, Chen,
  Goldie, Mirhoseini, McKinnon et~al.}]{bai2022constitutional}
Yuntao Bai, Saurav Kadavath, Sandipan Kundu, Amanda Askell, Jackson Kernion,
  Andy Jones, Anna Chen, Anna Goldie, Azalia Mirhoseini, Cameron McKinnon,
  et~al. 2022.
\newblock Constitutional ai: Harmlessness from ai feedback.
\newblock \emph{arXiv preprint arXiv:2212.08073}.

\bibitem[{Bhalotia(2022)}]{yc-startups}
Akshay Bhalotia. 2022.
\newblock Yc company scraper.
\newblock \url{https://github.com/akshaybhalotia/yc_company_scraper}.

\bibitem[{Bills et~al.(2023)Bills, Cammarata, Mossing, Tillman, Gao, Goh,
  Sutskever, Leike, Wu, and
  Saunders}]{Bills_Cammarata_Mossing_Tillman_Gao_Goh_Sutskever_Leike_Wu_Saunders_2023}
Steven Bills, Nick Cammarata, Dan Mossing, Henk Tillman, Leo Gao, Gabriel Goh,
  Ilya Sutskever, Jan Leike, Jeff Wu, and William Saunders. 2023.
\newblock \href
  {https://openai.com/research/language-models-can-explain-neurons-in-language-models}
  {[link]}.

\bibitem[{Blei et~al.(2003)Blei, Ng, and Jordan}]{blei2003latent}
David~M Blei, Andrew~Y Ng, and Michael~I Jordan. 2003.
\newblock Latent dirichlet allocation.
\newblock \emph{Journal of machine Learning research}, 3(Jan):993--1022.

\bibitem[{Cao et~al.(2021)Cao, Brbic, and Leskovec}]{cao2021open}
Kaidi Cao, Maria Brbic, and Jure Leskovec. 2021.
\newblock Open-world semi-supervised learning.
\newblock \emph{arXiv preprint arXiv:2102.03526}.

\bibitem[{Carmel et~al.(2009)Carmel, Roitman, and
  Zwerdling}]{carmel2009enhancing}
David Carmel, Haggai Roitman, and Naama Zwerdling. 2009.
\newblock Enhancing cluster labeling using wikipedia.
\newblock In \emph{Proceedings of the 32nd international ACM SIGIR conference
  on Research and development in information retrieval}, pages 139--146.

\bibitem[{Chang et~al.(2009)Chang, Gerrish, Wang, Boyd-Graber, and
  Blei}]{chang2009reading}
Jonathan Chang, Sean Gerrish, Chong Wang, Jordan Boyd-Graber, and David Blei.
  2009.
\newblock Reading tea leaves: How humans interpret topic models.
\newblock \emph{Advances in neural information processing systems}, 22.

\bibitem[{Chung et~al.(2022)Chung, Hou, Longpre, Zoph, Tay, Fedus, Li, Wang,
  Dehghani, Brahma et~al.}]{chung2022scaling}
Hyung~Won Chung, Le~Hou, Shayne Longpre, Barret Zoph, Yi~Tay, William Fedus,
  Eric Li, Xuezhi Wang, Mostafa Dehghani, Siddhartha Brahma, et~al. 2022.
\newblock Scaling instruction-finetuned language models.
\newblock \emph{arXiv preprint arXiv:2210.11416}.

\bibitem[{Downey et~al.(2015)Downey, Bhagavatula, and
  Yang}]{downey-etal-2015-efficient}
Doug Downey, Chandra Bhagavatula, and Yi~Yang. 2015.
\newblock \href {https://doi.org/10.3115/v1/P15-1075} {Efficient methods for
  inferring large sparse topic hierarchies}.
\newblock In \emph{Proceedings of the 53rd Annual Meeting of the Association
  for Computational Linguistics and the 7th International Joint Conference on
  Natural Language Processing (Volume 1: Long Papers)}, pages 774--784,
  Beijing, China. Association for Computational Linguistics.

\bibitem[{Eyuboglu et~al.(2022)Eyuboglu, Varma, Saab, Delbrouck, Lee-Messer,
  Dunnmon, Zou, and R{\'e}}]{eyuboglu2022domino}
Sabri Eyuboglu, Maya Varma, Khaled Saab, Jean-Benoit Delbrouck, Christopher
  Lee-Messer, Jared Dunnmon, James Zou, and Christopher R{\'e}. 2022.
\newblock Domino: Discovering systematic errors with cross-modal embeddings.
\newblock \emph{arXiv preprint arXiv:2203.14960}.

\bibitem[{Gilardi et~al.(2023)Gilardi, Alizadeh, and
  Kubli}]{gilardi2023chatgpt}
Fabrizio Gilardi, Meysam Alizadeh, and Ma{\"e}l Kubli. 2023.
\newblock Chatgpt outperforms crowd-workers for text-annotation tasks.
\newblock \emph{arXiv preprint arXiv:2303.15056}.

\bibitem[{Habernal and Gurevych(2016)}]{habernal-gurevych-2016-makes}
Ivan Habernal and Iryna Gurevych. 2016.
\newblock \href {https://doi.org/10.18653/v1/D16-1129} {What makes a convincing
  argument? empirical analysis and detecting attributes of convincingness in
  web argumentation}.
\newblock In \emph{Proceedings of the 2016 Conference on Empirical Methods in
  Natural Language Processing}, pages 1214--1223, Austin, Texas. Association
  for Computational Linguistics.

\bibitem[{He(2020)}]{rate-my-prof}
Jibo He. 2020.
\newblock \href {https://data.mendeley.com/datasets/fvtfjyvw7d/2} {{Big Data
  Set from RateMyProfessor.com for Professors' Teaching Evaluation}}.

\bibitem[{He and McAuley(2016)}]{he2016ups}
Ruining He and Julian McAuley. 2016.
\newblock Ups and downs: Modeling the visual evolution of fashion trends with
  one-class collaborative filtering.
\newblock In \emph{proceedings of the 25th international conference on world
  wide web}, pages 507--517.

\bibitem[{Hu et~al.(2014)Hu, Boyd-Graber, Satinoff, and
  Smith}]{hu2014interactive}
Yuening Hu, Jordan Boyd-Graber, Brianna Satinoff, and Alison Smith. 2014.
\newblock Interactive topic modeling.
\newblock \emph{Machine learning}, 95:423--469.

\bibitem[{Krizhevsky et~al.(2009)Krizhevsky, Hinton
  et~al.}]{krizhevsky2009learning}
Alex Krizhevsky, Geoffrey Hinton, et~al. 2009.
\newblock Learning multiple layers of features from tiny images.

\bibitem[{Kulkarni(2018)}]{abc-headlines}
Rohit Kulkarni. 2018.
\newblock \href {https://doi.org/10.7910/DVN/SYBGZL} {{A Million News
  Headlines}}.

\bibitem[{Kulkarni(2020)}]{clickbait-headlines}
Rohit Kulkarni. 2020.
\newblock \href {https://www.kaggle.com/datasets/therohk/examine-the-examiner}
  {{The Examiner - Spam Clickbait Catalog}}.

\bibitem[{Lange et~al.(2004)Lange, Roth, Braun, and
  Buhmann}]{lange2004stability}
Tilman Lange, Volker Roth, Mikio~L Braun, and Joachim~M Buhmann. 2004.
\newblock Stability-based validation of clustering solutions.
\newblock \emph{Neural computation}, 16(6):1299--1323.

\bibitem[{Luu et~al.(2014)Luu, Kim, and Ng}]{luu-etal-2014-taxonomy}
Anh~Tuan Luu, Jung-jae Kim, and See~Kiong Ng. 2014.
\newblock \href {https://doi.org/10.3115/v1/D14-1088} {Taxonomy construction
  using syntactic contextual evidence}.
\newblock In \emph{Proceedings of the 2014 Conference on Empirical Methods in
  Natural Language Processing ({EMNLP})}, pages 810--819, Doha, Qatar.
  Association for Computational Linguistics.

\bibitem[{Mekala and Shang(2020)}]{mekala2020contextualized}
Dheeraj Mekala and Jingbo Shang. 2020.
\newblock Contextualized weak supervision for text classification.
\newblock In \emph{Proceedings of the 58th Annual Meeting of the Association
  for Computational Linguistics}, pages 323--333.

\bibitem[{Meng et~al.(2020)Meng, Huang, Wang, Wang, Zhang, Zhang, and
  Han}]{meng2020discriminative}
Yu~Meng, Jiaxin Huang, Guangyuan Wang, Zihan Wang, Chao Zhang, Yu~Zhang, and
  Jiawei Han. 2020.
\newblock Discriminative topic mining via category-name guided text embedding.
\newblock In \emph{Proceedings of The Web Conference 2020}, pages 2121--2132.

\bibitem[{Mitchell et~al.(2011)Mitchell, O'Sullivan, and
  Dunning}]{Mitchell2011PuLPA}
Stuart Mitchell, Michael~J. O'Sullivan, and Iain Dunning. 2011.
\newblock Pulp : A linear programming toolkit for python.

\bibitem[{Ni et~al.(2019)Ni, Li, and McAuley}]{amazon-reviews}
Jianmo Ni, Jiacheng Li, and Julian McAuley. 2019.
\newblock Justifying recommendations using distantly-labeled reviews and
  fine-grained aspects.
\newblock In \emph{Proceedings of the 2019 conference on empirical methods in
  natural language processing and the 9th international joint conference on
  natural language processing (EMNLP-IJCNLP)}, pages 188--197.

\bibitem[{OpenAI(2023)}]{openai2023gpt4}
OpenAI. 2023.
\newblock \href {http://arxiv.org/abs/2303.08774} {Gpt-4 technical report}.

\bibitem[{Pedregosa et~al.(2011)Pedregosa, Varoquaux, Gramfort, Michel,
  Thirion, Grisel, Blondel, Prettenhofer, Weiss, Dubourg, Vanderplas, Passos,
  Cournapeau, Brucher, Perrot, and Duchesnay}]{scikit-learn}
F.~Pedregosa, G.~Varoquaux, A.~Gramfort, V.~Michel, B.~Thirion, O.~Grisel,
  M.~Blondel, P.~Prettenhofer, R.~Weiss, V.~Dubourg, J.~Vanderplas, A.~Passos,
  D.~Cournapeau, M.~Brucher, M.~Perrot, and E.~Duchesnay. 2011.
\newblock Scikit-learn: Machine learning in {P}ython.
\newblock \emph{Journal of Machine Learning Research}, 12:2825--2830.

\bibitem[{Raffel et~al.(2020)Raffel, Shazeer, Roberts, Lee, Narang, Matena,
  Zhou, Li, and Liu}]{raffel2020exploring}
Colin Raffel, Noam Shazeer, Adam Roberts, Katherine Lee, Sharan Narang, Michael
  Matena, Yanqi Zhou, Wei Li, and Peter~J Liu. 2020.
\newblock Exploring the limits of transfer learning with a unified text-to-text
  transformer.
\newblock \emph{The Journal of Machine Learning Research}, 21(1):5485--5551.

\bibitem[{Scheurer et~al.(2023)Scheurer, Campos, Korbak, Chan, Chen, Cho, and
  Perez}]{scheurer2023training}
J{\'e}r{\'e}my Scheurer, Jon~Ander Campos, Tomasz Korbak, Jun~Shern Chan,
  Angelica Chen, Kyunghyun Cho, and Ethan Perez. 2023.
\newblock Training language models with language feedback at scale.
\newblock \emph{arXiv preprint arXiv:2303.16755}.

\bibitem[{Shang et~al.(2018)Shang, Liu, Jiang, Ren, Voss, and
  Han}]{shang2018automated}
Jingbo Shang, Jialu Liu, Meng Jiang, Xiang Ren, Clare~R Voss, and Jiawei Han.
  2018.
\newblock Automated phrase mining from massive text corpora.
\newblock \emph{IEEE Transactions on Knowledge and Data Engineering},
  30(10):1825--1837.

\bibitem[{Shang et~al.(2020)Shang, Zhang, Liu, Li, and Han}]{shang2020nettaxo}
Jingbo Shang, Xinyang Zhang, Liyuan Liu, Sha Li, and Jiawei Han. 2020.
\newblock Nettaxo: Automated topic taxonomy construction from text-rich
  network.
\newblock In \emph{Proceedings of The Web Conference 2020}, pages 1908--1919.

\bibitem[{Shu et~al.(2018)Shu, Xu, and Liu}]{shu2018unseen}
Lei Shu, Hu~Xu, and Bing Liu. 2018.
\newblock Unseen class discovery in open-world classification.
\newblock \emph{arXiv preprint arXiv:1801.05609}.

\bibitem[{Singh et~al.(2023)Singh, Hsu, Antonello, Jain, Huth, Yu, and
  Gao}]{singh2023explaining}
Chandan Singh, Aliyah~R Hsu, Richard Antonello, Shailee Jain, Alexander~G Huth,
  Bin Yu, and Jianfeng Gao. 2023.
\newblock Explaining black box text modules in natural language with language
  models.
\newblock \emph{arXiv preprint arXiv:2305.09863}.

\bibitem[{Singh et~al.(2022)Singh, Morris, Aneja, Rush, and
  Gao}]{singh2022explaining}
Chandan Singh, John~X Morris, Jyoti Aneja, Alexander~M Rush, and Jianfeng Gao.
  2022.
\newblock Explaining patterns in data with language models via interpretable
  autoprompting.
\newblock \emph{arXiv preprint arXiv:2210.01848}.

\bibitem[{Su et~al.(2022)Su, Kasai, Wang, Hu, Ostendorf, Yih, Smith,
  Zettlemoyer, Yu et~al.}]{su2022one}
Hongjin Su, Jungo Kasai, Yizhong Wang, Yushi Hu, Mari Ostendorf, Wen-tau Yih,
  Noah~A Smith, Luke Zettlemoyer, Tao Yu, et~al. 2022.
\newblock One embedder, any task: Instruction-finetuned text embeddings.
\newblock \emph{arXiv preprint arXiv:2212.09741}.

\bibitem[{Thomason et~al.(2016)Thomason, Sinapov, Svetlik, Stone, and
  Mooney}]{thomason2016learning}
Jesse Thomason, Jivko Sinapov, Maxwell Svetlik, Peter Stone, and Raymond~J
  Mooney. 2016.
\newblock Learning multi-modal grounded linguistic semantics by playing" i
  spy".
\newblock In \emph{IJCAI}, pages 3477--3483.

\bibitem[{Treeratpituk and Callan(2006)}]{treeratpituk2006automatically}
Pucktada Treeratpituk and Jamie Callan. 2006.
\newblock Automatically labeling hierarchical clusters.
\newblock In \emph{Proceedings of the 2006 international conference on Digital
  government research}, pages 167--176.

\bibitem[{Vaze et~al.(2022)Vaze, Han, Vedaldi, and
  Zisserman}]{vaze2022generalized}
Sagar Vaze, Kai Han, Andrea Vedaldi, and Andrew Zisserman. 2022.
\newblock Generalized category discovery.
\newblock In \emph{Proceedings of the IEEE/CVF Conference on Computer Vision
  and Pattern Recognition}, pages 7492--7501.

\bibitem[{Wang et~al.(2022)Wang, Yang, Huang, Jiao, Yang, Jiang, Majumder, and
  Wei}]{wang2022text}
Liang Wang, Nan Yang, Xiaolong Huang, Binxing Jiao, Linjun Yang, Daxin Jiang,
  Rangan Majumder, and Furu Wei. 2022.
\newblock Text embeddings by weakly-supervised contrastive pre-training.
\newblock \emph{arXiv preprint arXiv:2212.03533}.

\bibitem[{Wang et~al.(2023)Wang, Wang, Liu, and Shang}]{wang2023wot}
Tianle Wang, Zihan Wang, Weitang Liu, and Jingbo Shang. 2023.
\newblock Wot-class: Weakly supervised open-world text classification.
\newblock \emph{arXiv preprint arXiv:2305.12401}.

\bibitem[{Wolf et~al.(2020)Wolf, Debut, Sanh, Chaumond, Delangue, Moi, Cistac,
  Rault, Louf, Funtowicz et~al.}]{wolf2020transformers}
Thomas Wolf, Lysandre Debut, Victor Sanh, Julien Chaumond, Clement Delangue,
  Anthony Moi, Pierric Cistac, Tim Rault, R{\'e}mi Louf, Morgan Funtowicz,
  et~al. 2020.
\newblock Transformers: State-of-the-art natural language processing.
\newblock In \emph{Proceedings of the 2020 conference on empirical methods in
  natural language processing: system demonstrations}, pages 38--45.

\bibitem[{Xie et~al.(2023)Xie, Luo, Li, Liu, Shen, and Shou}]{xie2023open}
Jinheng Xie, Zhaochuan Luo, Yuexiang Li, Haozhe Liu, Linlin Shen, and
  Mike~Zheng Shou. 2023.
\newblock Open-world weakly-supervised object localization.
\newblock \emph{arXiv preprint arXiv:2304.08271}.

\bibitem[{Zhang et~al.(2018)Zhang, Tao, Chen, Shen, Jiang, Sadler, Vanni, and
  Han}]{zhang2018taxogen}
Chao Zhang, Fangbo Tao, Xiusi Chen, Jiaming Shen, Meng Jiang, Brian Sadler,
  Michelle Vanni, and Jiawei Han. 2018.
\newblock Taxogen: Unsupervised topic taxonomy construction by adaptive term
  embedding and clustering.
\newblock In \emph{Proceedings of the 24th ACM SIGKDD International Conference
  on Knowledge Discovery \& Data Mining}, pages 2701--2709.

\bibitem[{Zhang et~al.(2015)Zhang, Zhao, and LeCun}]{zhang2015character}
Xiang Zhang, Junbo Zhao, and Yann LeCun. 2015.
\newblock Character-level convolutional networks for text classification.
\newblock \emph{Advances in neural information processing systems}, 28.

\bibitem[{Zhao and Mac~Aodha(2023)}]{zhao2023incremental}
Bingchen Zhao and Oisin Mac~Aodha. 2023.
\newblock Incremental generalized category discovery.
\newblock \emph{arXiv preprint arXiv:2304.14310}.

\bibitem[{Zheng et~al.(2022)Zheng, Li, Hong, Petersson, and
  Barnes}]{zheng2022towards}
Jiyang Zheng, Weihao Li, Jie Hong, Lars Petersson, and Nick Barnes. 2022.
\newblock Towards open-set object detection and discovery.
\newblock In \emph{Proceedings of the IEEE/CVF Conference on Computer Vision
  and Pattern Recognition}, pages 3961--3970.

\bibitem[{Zhong et~al.(2022)Zhong, Snell, Klein, and
  Steinhardt}]{zhong2022describing}
Ruiqi Zhong, Charlie Snell, Dan Klein, and Jacob Steinhardt. 2022.
\newblock Describing differences between text distributions with natural
  language.
\newblock In \emph{International Conference on Machine Learning}, pages
  27099--27116. PMLR.

\bibitem[{Zhong et~al.(2023)Zhong, Zhang, Li, Ahn, Klein, and
  Steinhardt}]{zhong2023goal}
Ruiqi Zhong, Peter Zhang, Steve Li, Jinwoo Ahn, Dan Klein, and Jacob
  Steinhardt. 2023.
\newblock Goal driven discovery of distributional differences via language
  descriptions.
\newblock \emph{arXiv preprint arXiv:2302.14233}.

\bibitem[{Zhu et~al.(2022)Zhu, Liang, and Zou}]{zhu2022gsclip}
Zhiying Zhu, Weixin Liang, and James Zou. 2022.
\newblock Gsclip: A framework for explaining distribution shifts in natural
  language.
\newblock \emph{arXiv preprint arXiv:2206.15007}.

\end{thebibliography}
\bibliographystyle{acl_natbib}

\appendix

\section{Prompt Templates} \label{app:prompt-templates}

\begin{figure*}
    \centering
    \includegraphics[width=0.8\linewidth]{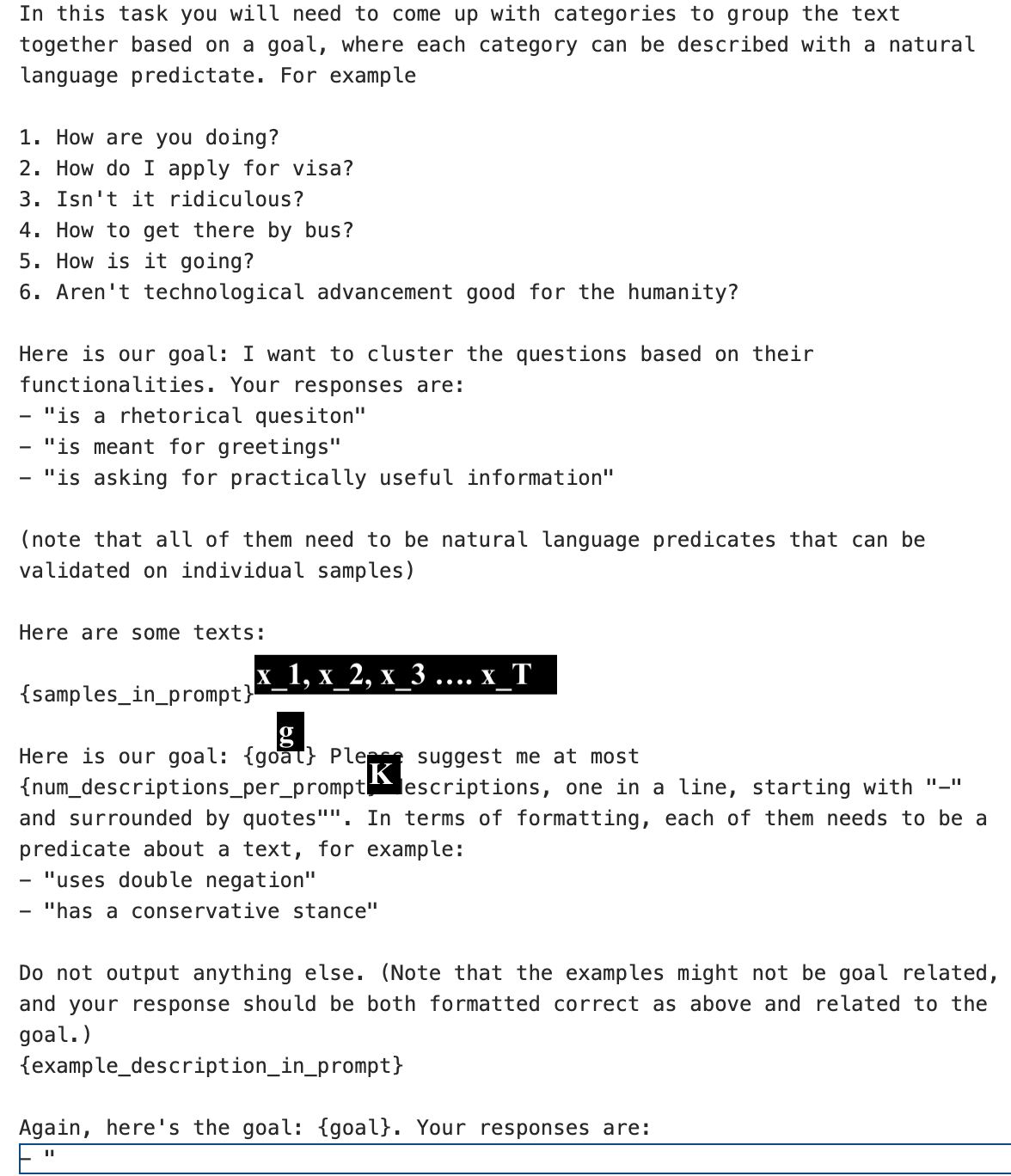}
    \caption{The template we used to propose candidate explanations, where we will substitute the corresponding variables to construct the prompt. }
    \label{fig:proposal-simple}
\end{figure*}

\noindent \textbf{Proposal Stage.} 
Figure \ref{fig:proposal-simple} shows the prompt we used on \syndataset to propose simple explanations for the clusters.
Figure~\ref{fig:perturbed-proposal-simple} shows the perturbed prompt for conducting prompt sensitivity analysis.
Figure \ref{fig:proposal-complex} shows the formatting instruction we used on \opendataset to propose more sophisticated explanations for the clusters. 

\begin{figure*}
        \centering
    \includegraphics[width=0.8\linewidth]{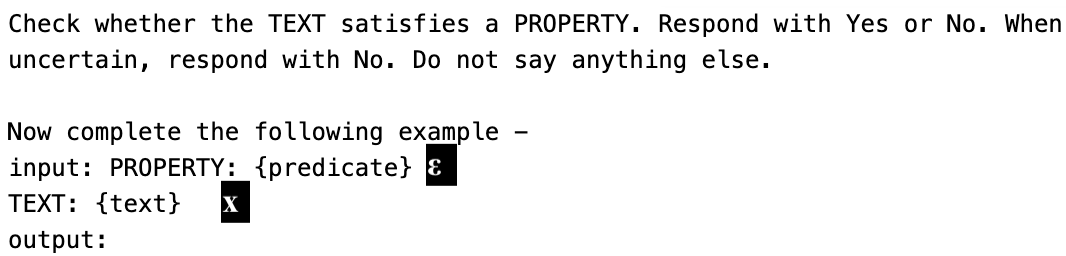}
    \caption{The template we used for the assigner which decides whether a candidate explanation supports a text sample.}
    \label{fig:assigner}
\end{figure*}

\begin{figure*}
    \centering
    \includegraphics[width=0.8\linewidth]{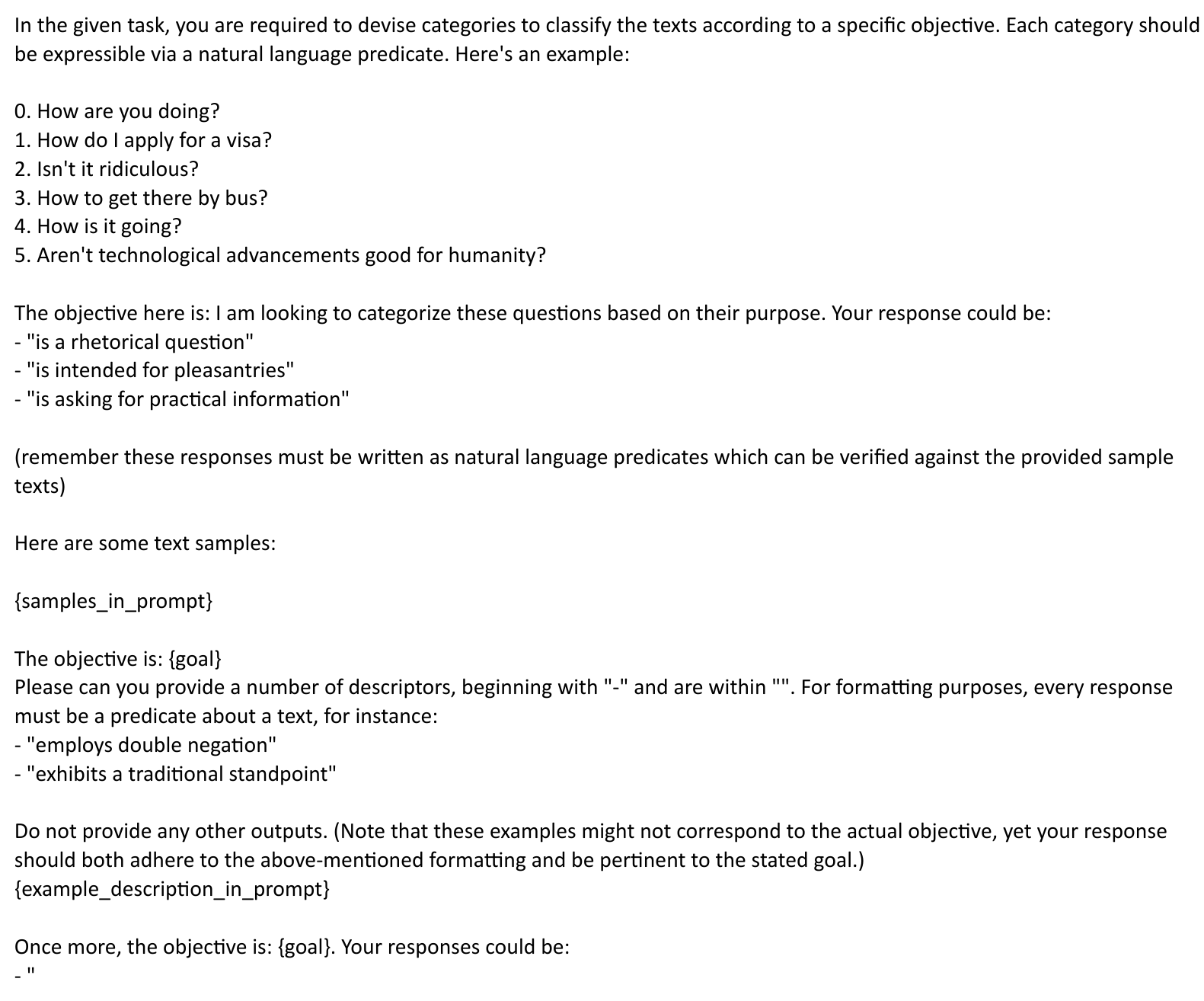}
    \caption{The perturbed proposal template.}
    \label{fig:perturbed-proposal-simple}
\end{figure*}

\begin{figure*}
    \centering
    \includegraphics[width=0.8\linewidth]{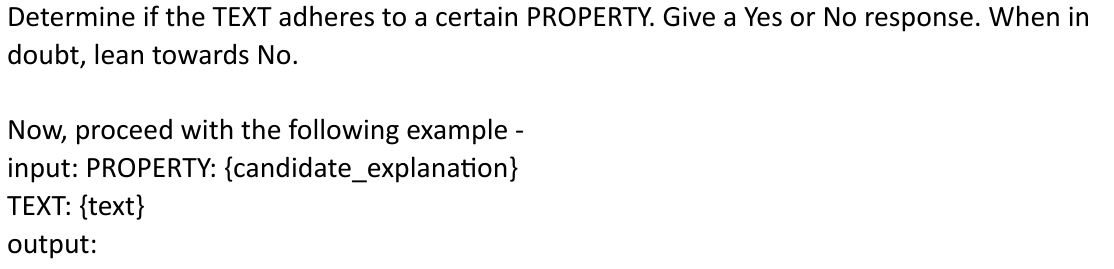}
    \caption{The perturbed assigner template.}
    \label{fig:perturbed-assigner-simple}
\end{figure*}

\begin{figure*}
    \centering
    \includegraphics[width=0.8\linewidth]{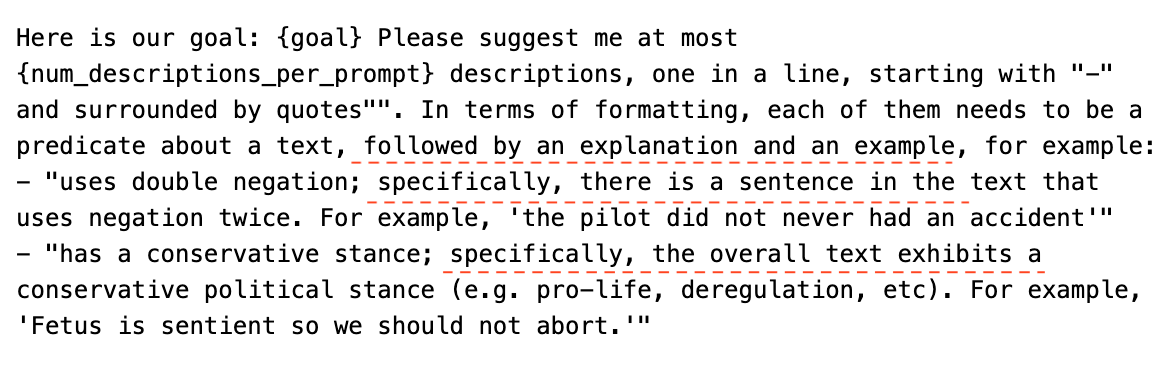}
    \caption{We used the same template to propose hypotheses in Section \ref{sec:open-ended}, except that we changed the formatting instruction to propose more sophisticated predicates. The changed part is shown above and the key changes are underlined in red.}
    \label{fig:proposal-complex}
\end{figure*}

\noindent \textbf{Assignment Stage.} 
Figure \ref{fig:assigner} shows the prompt we used to check whether an explanation supports a sample.
Figure~\ref{fig:perturbed-assigner-simple} shows the perturbed prompt for conducting prompt sensitivity analysis.

\noindent \textbf{Prompt to Commit to a Single Explanation.}

\begin{quote}
    \textit{Predicate 0:} $\epsilon_{1}$ \\
    \textit{Predicate 1:} $\epsilon_{2}$ \\
    \dots \\
    \textit{Predicate K:} $\epsilon_{K}$ \\
    \textit{Text: }$x$. \textit{\\Choose the Predicate the matches the Text the most.}
\end{quote}

\section{Selection Stage Implementation} \label{app:selection-pseudo}
\begin{figure*}
    \centering
    \includegraphics[width=0.8\linewidth]{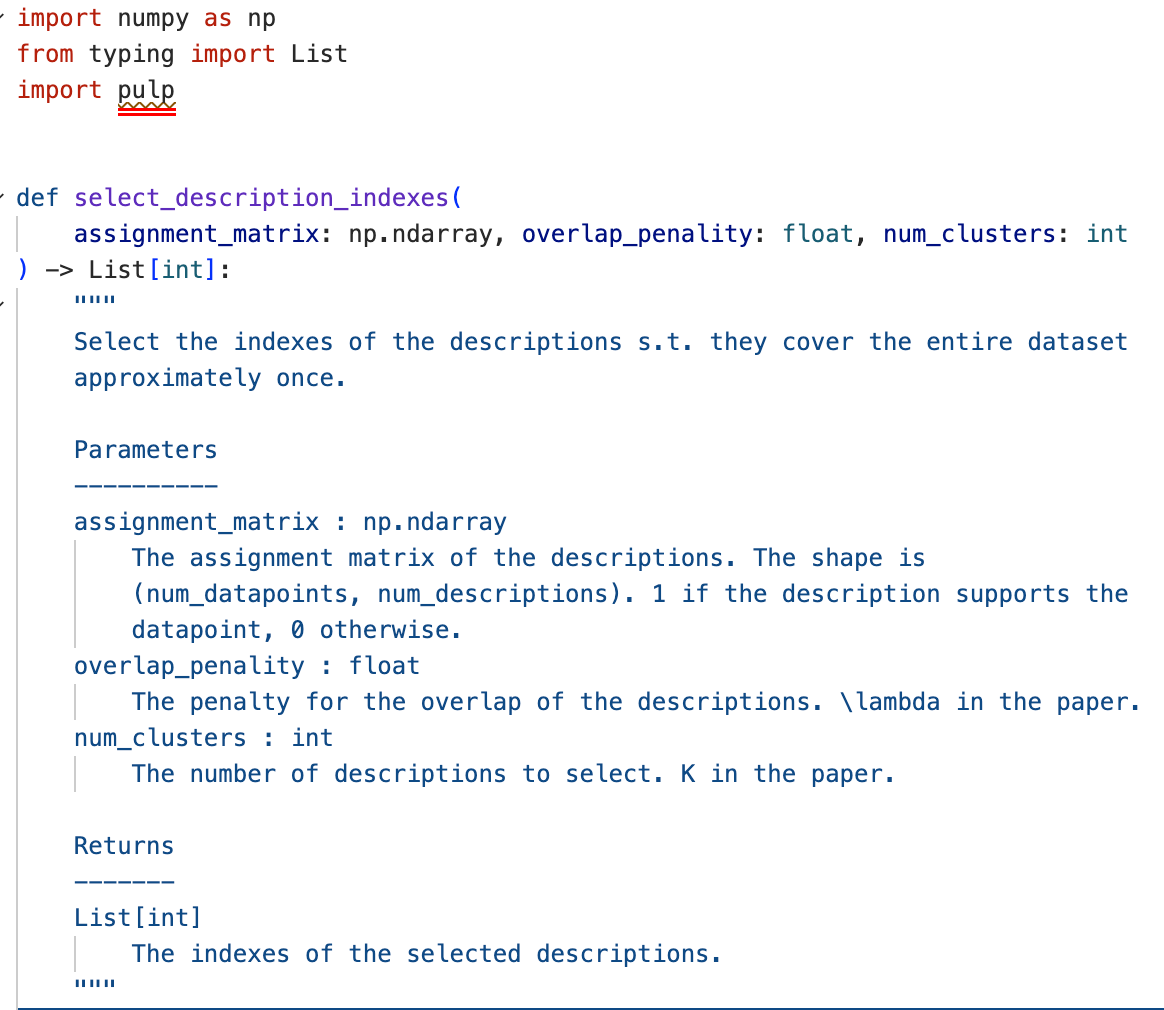}
    \caption{The function signature of the selection stage. The function body can be seen in Figure \ref{fig:selection-sig}}
    \label{fig:selection-sig}
\end{figure*}

\begin{figure*}
    \centering
    \includegraphics[width=0.8\linewidth]{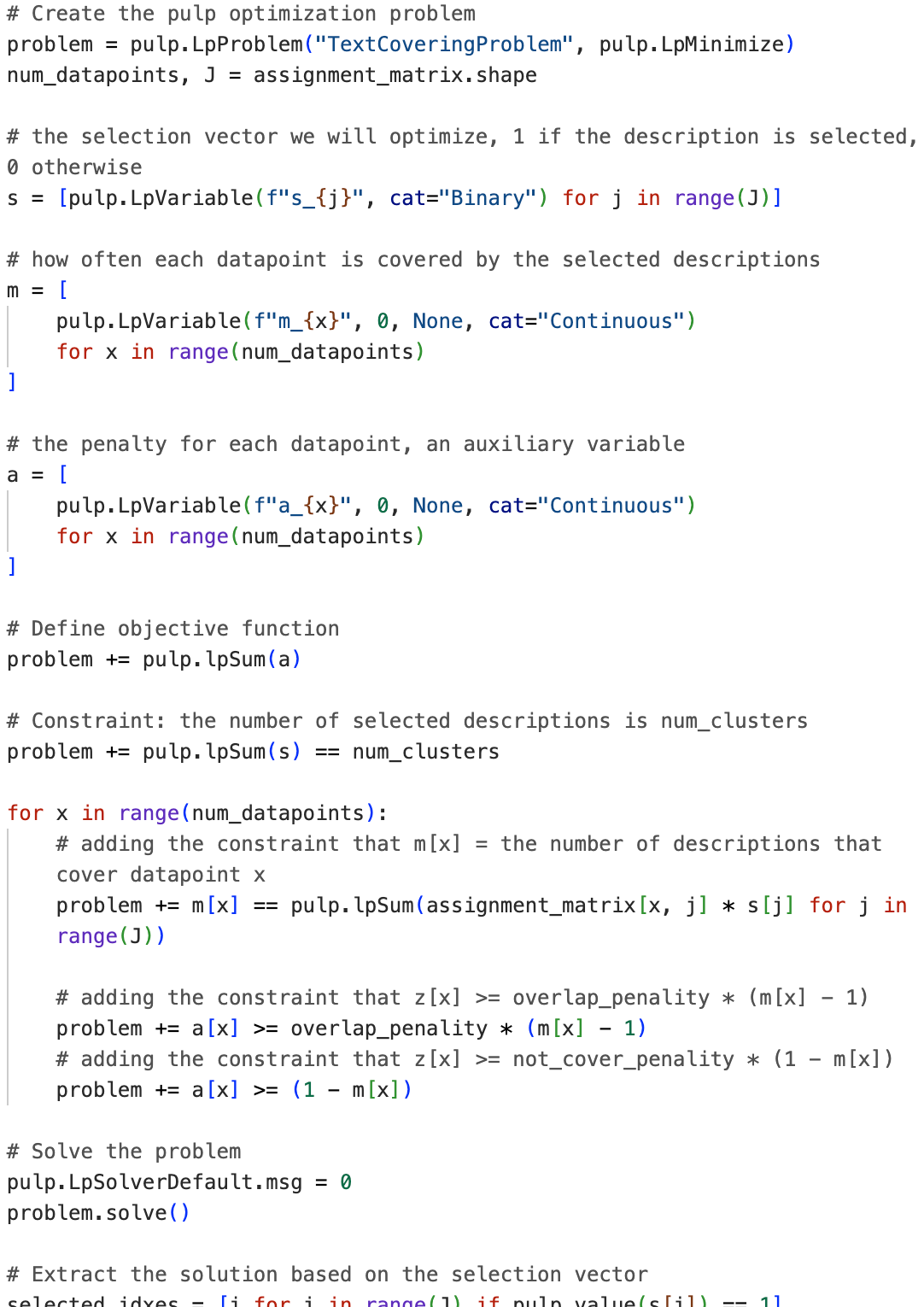}
    \caption{The function body of the selection stage. The function signature can be seen in Figure \ref{fig:selection-sig}}
    \label{fig:selection-body}
\end{figure*}

To help the reader understand our algorithm for selecting the descriptions, we include our python implementation in Figure \ref{fig:selection-sig} and \ref{fig:selection-body}. 

\section{Synthesizing \syndataset{}} \label{app:syngoal}

We first describe the four values for each dimension and then discuss the prompts we used to generate the text samples in \syndataset{}. 

\noindent \textbf{\underline{T}opic}: 1.``\textit{has a topic of what sports to do to improve your health}'', 2. ``\textit{has a topic of a new anime has been announced}'', 3. ``\textit{has a topic of a tech company releases a new groundbreaking paper}'', and 4. ``\textit{has a topic of how to improve your productivity}''.

\noindent \textbf{Writing \underline{S}tyle} 1. ``\textit{has a writing style of twitter}'', 2. ``\textit{has a writing style of screen play}'', 3. ``\textit{has a writing style of rap}'', and 4) ``\textit{has a writing style of poem}''.

\noindent \textbf{\underline{L}anguage} 1. ``\textit{has a natural language of English}'', 2. ``\textit{has a natural language of French}'', 3. ``\textit{has a natural language of Deutsch}'', and 4) ``\textit{has a natural language of Spanish}''.

To generate the text samples conditioned on three dimensions, we first generated ``content samples'' based on the topic. 
For each topic, we first asked GPT-4 \citet{openai2023gpt4} to generate 40 English news summary for each topic using as diverse vocabulary as possible.
Here is the prompt template we used, where we substituted \{topic\} with the topic we want to condition on. 

\begin{quote}
   ``\textit{Write 40 news 2-sentence paragraphs about topic \{topic\}, using as diverse vocabulary as possible. We prefer being concrete; e.g. we prefer 'James Lebron joins Lake to strengthen the team; how likely is the team going to win next time?', rather than genric statements ``it's common to recruit new team members''. Additionally, you cannot use words that are directly used in the topic.\\\\
    The list continues below.}''
\end{quote}

For each of the 64 value combinations, we first sampled 16 text samples from the content samples based on its topic, and then for then prompted \texttt{Claude-1.3} ~\citep{bai2022constitutional} to rewrite it with a different style and language.
The template we used is as follows, where we substituted \{original\_text\} with the text to be rewritten, and \{style\} and \{language\} to be conditioned on:

\begin{quote}
    ``\textit{\{original\_text\}\\\\
    Rewrite the above paragraph in the style of \{style\} in \{language\}}.''
\end{quote}

\section{Other Techniques Used by \methodname{}} \label{sec:other}

\noindent \textbf{Multiple Iterations of \methodname{}.}
As mentioned in Section \ref{sec:propose}, since the proposer's context window might not be long enough to contain the entire corpus, the proposer might never ``see'' some samples or their similar variants;
consequently, some samples might not be supported by any of the proposed explanations.
Therefore, after one iteration of \methodname{}, we collect all samples not supported by any explanation and use them to propose candidate explanations again, hoping that some of the new candidates will support them.
To ensure broad supports over the entire corpus, we ran \methodname{} for 5 iterations. 

\noindent \textbf{Commit to a Single Cluster.}
At the end of \methodname{}, some samples might be supported by multiple selected explanations.
However, the user or a benchmark might require each sample to commit to one single most appropriate cluster.
Consider the following sample in a news corpus: $x$=``\dots  \textit{after years of disputes over security and national pride, Germany and the United\_States signed agreed to build a new American embassy on the empty lot \dots}''. $x$ is supported by both explanations $\epsilon_{1}$=``\textit{is related to real estate development}'' and $\epsilon_{2}$=``\textit{is related to international politics}''. 
While both $\epsilon_{1}$ and $\epsilon_{2}$ literally support $x$, $\epsilon_{2}$ is more appropriate.
To commit each sample to one single cluster, we prompted an LM with all selected explanations and asked it to choose one of them as our final commitment.
See our prompt template in Appendix \ref{app:prompt-templates}.

\section{Potential Effects of Memorization} \label{app:memorization}
The leakage of test data might affect both the proposer \texttt{gpt-3.5-turbo} and the assigner \texttt{flan-t5-xl} based on T5 \cite{raffel2020exploring}.

For the proposer, since the prompt represents a novel usage of large language model, we do not expect the proposer prompt specifically based on (AG), (DB), and (NYT) to have appeared in the pre-training corpus. 
That said, it is plausible that there are similar texts which produce categories based on other corpus, and they might have improved LM's capability to perform ``in-context clustering'';
on the other hand, however, such a consideration is irrelevant since we are not claiming that \texttt{gpt-3.5-turbo} can perform ``in-context clustering'' zero-shot without any similar training data.
Finally, \methodname{} recovers two of the uncommon topic labels (``\textit{anime}'' and ``\textit{productivity}'') for (SYN), produces reasonable explanations not identical to the reference on (DB), and generates novel explanations for the clusters in Section \ref{sec:open-ended}; these empirical evidence suggests that the capability of the proposer is largely not due to memorizing the training data. 

As for the assigner, it has a similar functionality as classification or entailment. Since both of these tasks are already relatively straightforward for state-of-the-art language models ~\citep{gilardi2023chatgpt}, we do not consider the potential effect of memorization on (NYT), (DB), and (AG) to play a significant role in our evaluation.

\section{Per-stage Evaluations} \label{app:per-stage}
In addition to an automatic end-to-end evaluation discussed in Section~\ref{sec:benchmark}, we also present an automatic \textit{per-stage} evaluation to better understand the quality of each stage of the \methodname{} algorithm. 

\noindent\textbf{Assign Stage.}
We first evaluate the second Assign stage, where we try to understand how well the assigner can recover the clusters when given the reference explanations of each cluster. 
This evaluation will measure the unavoidable discrepancy that might be brought by an imperfect assigner, such that later evaluations for other stages can marginalize out this discrepancy.
We formalize the evaluation as following:
Recall that we denote the corpus as $X$. Let $\overline{C_k}$ be the set of texts that belong to class k and $C_k$ the set of texts that are supported by the reference explanations of the class. We define the recall as
\begin{equation} \label{eq:step-recall}
    \frac{|\overline{C_k} \cap C_k|}{|\overline{C_k}|}
\end{equation}
and the specificity as 
\begin{equation} \label{eq:step-specificity}
    \frac{|X - \overline{C_k} \cap X - C_k|}{|X - \overline{C_k}|}
\end{equation}
Then, we take the average of recall and specificity as the score for this class and the average score over all classes as the score for the assigner on the dataset. 
The reason for analyzing recall and specificity is that recall is invariant if we found a superset of a class (e.g., a parent class in a hierarchy) and specificity is invariant when we find a subset of a class (e.g., a child class in a hierarchy). For the assigner, we expect the exact clusters to be found, hence we consider both recall and specificity. In latter two evaluations, since our method can operate in a top-down hierarchical manner, we focus on the recall, but not the specificity.

\begin{table}[]
\small
    \centering
\begin{tabular}{lll}
\toprule
                              reference explanation &  recall &  specificity \\
\midrule
                ``\textit{\phantom{} Company}'' & 44 &  100 \\
``\textit{\phantom{} Educational Institution}'' & 63 &  100 \\
                 ``\textit{\phantom{} Artist}'' & 26 &  100 \\
                ``\textit{\phantom{} Athlete}'' & 99 &  100 \\
          ``\textit{\phantom{} Office Holder}'' & 92 &  100 \\
 ``\textit{\phantom{} Mean Of Transportation}'' & 64 &  100 \\
               ``\textit{\phantom{} Building}'' & 35 &  100 \\
          ``\textit{\phantom{} Natural Place}'' & 41 &  97 \\
                ``\textit{\phantom{} Village}'' & 87 &  100 \\
                 ``\textit{\phantom{} Animal}'' & 76 &  100 \\
                  ``\textit{\phantom{} Plant}'' & 91 &  100 \\
                  ``\textit{\phantom{} Album}'' & 46 &  100 \\
                   ``\textit{\phantom{} Film}'' & 64 &  100 \\
           ``\textit{\phantom{} Written Work}'' & 28 &  100 \\
           \midrule
           \midrule
           \textbf{Assigner Score} & \multicolumn{2}{c}{80} \\
\bottomrule
\end{tabular}
    \caption{Assign-stage evaluation on (DB)pedia of \texttt{flan-t5}.  We abbreviate each explanation by removing the prefix ``\textit{has a topic of}'' (e.g., ``\textit{\phantom{} Company}'' corresponds to a full explanation ``\textit{\phantom{} has a topic of Company.}''). }
    \label{tab:app-eval-assign-ex}
\end{table}
\begin{table}[]
    \centering
    \begin{tabular}{lc}
    \toprule
    Dataset & Assign Score \\
    \midrule
    (DB)     &  80 \\
    (NYT)    &  73 \\
    \bottomrule
    \end{tabular}
    \caption{Assign-stage evaluation on (DB)pedia and (NYT) Topics of \texttt{flan-t5}.}
    \label{tab:app-eval-assign}
\end{table}

In Table~\ref{tab:app-eval-assign}, we illustrate the results for the assigner (and the later evaluations) on the two harder datasets ((NYT) Topics and (DB)pedia) where our method achieves a reasonable, but not perfect performance. Table~\ref{tab:app-eval-assign-ex} contains a per-explanation example for (DB)pedia.

\noindent\textbf{Propose Stage.}
In the Propose Stage, we would like to understand how well the proposed explanations capture the reference explanations. Therefore, for each reference explanation, we look into the proposed description that has highest recall to the reference explanation, as our method can operate in a top-down hierarchical manner to identify more specific clusters. We average the highest recall for each reference explanation as the score for the proposer.

\begin{table}[]
\small
    \resizebox{\linewidth}{!}{
    \centering
\begin{tabular}{lll}
\toprule
                    reference explanation &  proposed explanation & recall  \\
\midrule
                ``\textit{\phantom{} Company}'' & ``\textit{\phantom{} history}'' & 46  \\
``\textit{\phantom{} Educational Institution}'' & ``\textit{\phantom{} history}'' & 56  \\
                 ``\textit{\phantom{} Artist}'' & ``\textit{\phantom{} entertainment}'' & 79  \\
                ``\textit{\phantom{} Athlete}'' & ``\textit{\phantom{} sports}'' & 99  \\
          ``\textit{\phantom{} Office Holder}'' & ``\textit{\phantom{} politics and ...}'' & 87  \\
 ``\textit{\phantom{} Mean Of Transportation}'' & ``\textit{\phantom{} history}'' & 74  \\
               ``\textit{\phantom{} Building}'' & ``\textit{\phantom{} history}'' & 69  \\
          ``\textit{\phantom{} Natural Place}'' & ``\textit{\phantom{} geography}'' & 97  \\
                ``\textit{\phantom{} Village}'' & ``\textit{\phantom{} geography}'' & 98  \\
                 ``\textit{\phantom{} Animal}'' & ``\textit{\phantom{} zoology}'' & 80  \\
                  ``\textit{\phantom{} Plant}'' & ``\textit{\phantom{} botany}'' & 96  \\
                  ``\textit{\phantom{} Album}'' & ``\textit{\phantom{} entertainment}'' & 60  \\
                   ``\textit{\phantom{} Film}'' & ``\textit{\phantom{} entertainment}'' & 100  \\
           ``\textit{\phantom{} Written Work}'' & ``\textit{\phantom{} entertainment}'' & 76  \\
           \midrule
           \midrule
           \textbf{Proposer Score} & \multicolumn{2}{c}{80} \\
\bottomrule
\end{tabular}
}
    \caption{Propose-stage evaluation on (DB)pedia of \texttt{gpt-3.5-turbo}, the used reference assigner is \texttt{flan-t5}.}
    \label{tab:app-eval-propose-ex}
\end{table}
\begin{table}[]
    \centering
    \begin{tabular}{lcc}
    \toprule
    Dataset & Propose Score & \# matched proposes\\
    \midrule
    (DB)     &  80 & 7 \\
    (NYT)    &  86 & 9 \\
    \bottomrule
    \end{tabular}
    \caption{Propose-stage evaluation on the 14-class (DB)pedia and 9-class (NYT) Topics of \texttt{gpt-3.5-turbo}. The score and the \# of matched proposes are averaged over three runs. }
    \label{tab:app-eval-propose}
\end{table}
To remedy variances and get better understanding of the propose stage, we ask the proposer to make a large number of proposes (we used 128 proposes for both datasets) and repeat the experiment three times with different random seeds; the random seeds vary the text that is given to the proposer, therefore, could yield different proposed explanations from the proposer. 
The results are presented in Table~\ref{tab:app-eval-propose}.  We additionally show the number of proposed explanations that are actually matched (i.e., that are the highest recall for some reference explanations).
Table~\ref{tab:app-eval-propose-ex} contains a per-explanation example for (DB)pedia.
Overall, the proposer has a high coverage over the reference explanations. For (DB)pedia, the number of matched proposes is one half of the true number of classes, likely due to general explanations matched (e.g., ``\textit{\phantom{} Album}'' and ``\textit{\phantom{} Film}'' both are matched by ``\textit{\phantom{} entertainment}'').

\begin{table}[]
\small
    \resizebox{\linewidth}{!}{
    \centering
\begin{tabular}{lll}
\toprule
                    reference explanation &  proposed explanation & recall  \\
\midrule
                ``\textit{\phantom{} Company}'' & ``\textit{\phantom{} technology}'' & 31  \\
``\textit{\phantom{} Educational Institution}'' & ``\textit{\phantom{} language}'' & 18  \\
                 ``\textit{\phantom{} Artist}'' & ``\textit{\phantom{} music}'' & 38  \\
                ``\textit{\phantom{} Athlete}'' & ``\textit{\phantom{} sports}'' & 99  \\
          ``\textit{\phantom{} Office Holder}'' & ``\textit{\phantom{} politics}'' & 79  \\
 ``\textit{\phantom{} Mean Of Transportation}'' & ``\textit{\phantom{} technology}'' & 63  \\
               ``\textit{\phantom{} Building}'' & ``\textit{\phantom{} architecture}'' & 43  \\
          ``\textit{\phantom{} Natural Place}'' & ``\textit{\phantom{} botany}'' & 36  \\
                ``\textit{\phantom{} Village}'' & ``\textit{\phantom{} language}'' & 67  \\
                 ``\textit{\phantom{} Animal}'' & ``\textit{\phantom{} zoology}'' & 80  \\
                  ``\textit{\phantom{} Plant}'' & ``\textit{\phantom{} botany}'' & 96  \\
                  ``\textit{\phantom{} Album}'' & ``\textit{\phantom{} music}'' & 57  \\
                   ``\textit{\phantom{} Film}'' & ``\textit{\phantom{} film}'' & 97  \\
           ``\textit{\phantom{} Written Work}'' & ``\textit{\phantom{} literature}'' & 45  \\
           \midrule
           \midrule
           \textbf{Selector Score} & \multicolumn{2}{c}{61} \\
\bottomrule
\end{tabular}
}
    \caption{Select-stage evaluation on (DB)pedia of our selection ILP algorithm, the used proposer is \texttt{gpt-3.5-turbo}, and the reference assigner is \texttt{flan-t5}.}
    \label{tab:app-eval-select-ex}
\end{table}
\begin{table*}[]
    \centering
    \begin{tabular}{lccc}
    \toprule
    Dataset & Propose Score & Select Score & \# matched selected proposes\\
    \midrule
    (DB)     & 80 &  62 & 10 \\
    (NYT)    & 86 &  69 & 7 \\
    \bottomrule
    \end{tabular}
    \caption{Select-stage evaluation on the 14-class (DB)pedia and 9-class (NYT) Topics of our selection ILP algorithm. The score and the \# of proposes are averaged over three runs. The number of selected proposes is the same as the class number, as it is enforced in the algorithm.}
    \label{tab:app-eval-select}
\end{table*}

\noindent\textbf{Select Stage.} In the Select Stage, we are also interested in how well the proposed explanations cover the desired reference explanations. We again use the average highest recall as the score. In Table~\ref{tab:app-eval-select} we show the score for the select method in \methodname{}. We note that there is a drop in coverage (i.e., a drop from propose score to select score) during the select phase, even though it is potentially possible to pick exactly all the matched proposes in the proposer. This indicates a potential room of improvement for the select algorithm.

\begin{table*}[]
    \centering
    \resizebox{\linewidth}{!}{
    \begin{tabular}{llccc}
    \toprule
    Dataset & Method & Select Score & \# selected proposes & \# matched selected proposes\\
    \midrule
    (DB)  & ILP w/ \# Clusters Constraint                &  62  & 14 & 10 \\
    (DB)  & ILP w/ \# Clusters Penalty                   &  59  & 8  & 8 \\
    (NYT)  & ILP w/ \# Clusters Constraint               &  69  & 9  & 7 \\
    (NYT)  & ILP w/ \# Clusters Penalty                  &  67  & 7  & 7 \\
    \bottomrule
    \end{tabular}
    }
    \caption{Select-stage evaluation on (DB)pedia and (NYT) Topics comparing the ILP method in our paper where we enforce the number of selected clusters with a constraint and a variation where we inject a small cost to favor small number of clusters. The score and the \# of proposes are averaged over three runs.}
    \label{tab:app-eval-select-noclusnum}
\end{table*}
\noindent\textbf{Clustering w/o Cluster Number Constraints.} We would like to point out that our select algorithm does not have to enforce a number of clusters\footnote{The reason that we do this is for fair comparison with prior clustering methods.}. We could have remove the constraint of number of clusters and add an penalty proportional to the number of selected clusters in the objective. We conduct an initial experiment by changing the ILP objective to=
\begin{equation}
    \mathcal{L} = a\cdot \mathbf{1} + 10 * (s \cdot \mathbf{1}),
\end{equation}
and show the results in Table~\ref{tab:app-eval-select-noclusnum}. Notably, by not specifying the cluster number to select, our algorithm is able to pick a more compact set of clusters with almost similar coverage. 

\section{Further Ablations} \label{app:further-ablation}

We conducted two ablations for \underline{\methodname{}} to study the contribution of 1) proposing multiple iterations and 2) our selection algorithm.
For 1) we compared to only running \underline{\methodname{}} for 1 iteration, where the proposer ``sees'' much fewer samples;
for 2) we compared to the algorithm of greedily selecting the clusters to maximize coverage and a variant of ILP that sets $\lambda=0$ (not penalizing the overlaps).

We report the performance in Table~\ref{tab:ablation}.
Overall, running \underline{\methodname{}} for five iterations improves over one iteration and using an ILP algorithm with a positive $\lambda$ improves the performance.

\begin{table*}[t]
    \small
    \centering
    \begin{tabular}{l|ccccccc}
    \toprule
    \multirow{2}{*}{Macro F$_1$(\%)} & {(AG)} & {(DB)} &\multicolumn{2}{c}{(NYT)} & \multicolumn{3}{c}{(SYN)} \\
    & Topic & Topic & Topic & Location & Topic & Language & Style \\
    \midrule
    \methodname          & 87 & \textbf{71} & 70 & \textbf{76} & \textbf{98} & 97 & \textbf{31} \\
    \midrule
    Only 1 Iteration     & \textbf{88} & 60 & 48 & \textbf{77} & 95 & 97 & 29  \\
    Selection w/ $\lambda=0$ & 74 & \textbf{71} &  57 & 72 & 98 & \textbf{99} & 29 \\
    Selection w/ Greedy     & 70 & 65 & 55 & 68 & 92 & 98 & 28 \\
    \bottomrule
    \end{tabular}
    \caption{
    \methodname with ablations without the iterative proposing technique and with different selection algorithms described in Section \ref{sec:performance}.
    Overall, running multiple iterations and using ILP with $\lambda > 0$ are helpful.
    }
    \label{tab:ablation}
\end{table*}

\section{\methodname{}-Generated Descriptions}

We compare the \methodname{}-generated explanations to the reference explanations;
for each pair of generated and reference explanations, we also compute the F$_1$ score between the two generated and the reference cluster (100 if they are identical and 0 if they are disjoint). 
Generally we found that the generated explanations are semantically relevant or even equivalent to the references (Table \ref{tab:app-descriptions-agnews}, \ref{tab:app-descriptions-dbpedia}, \ref{tab:app-descriptions-nyt-topic}, \ref{tab:app-descriptions-nyt-location}, \ref{tab:app-descriptions-syn-language}, and \ref{tab:app-descriptions-syn-topic});
the only exception is when we used \texttt{gpt-3.5-turbo} as the proposer and \texttt{Flan-T5} as the assigner to cluster based on writing styles on \syndataset{} (Table \ref{tab:app-descriptions-syn-style}), but the problem alleviates when we use better models (\texttt{gpt-4} as proposer, \texttt{gpt-3.5-turbo} as the assigner, Table \ref{tab:app-descriptions-syn-style-gpt4}). 

\begin{table*}[]
    \centering
\begin{tabular}{lll}
\toprule
                    reference explanation &                                     output explanation &  F$_1$ \\
\midrule
  ``\textit{has a topic of Politics}'' & ``\textit{has a topic of politics and social issues}'' &  84 \\
    ``\textit{has a topic of Sports}'' &                     ``\textit{has a topic of sports}'' &  97 \\
  ``\textit{has a topic of Business}'' &                    ``\textit{has a topic of finance}'' &  81 \\
``\textit{has a topic of Technology}'' &                 ``\textit{has a topic of technology}'' &  85 \\
\bottomrule
\end{tabular}
    \caption{AG's News, clustering based on Topics, proposer=\texttt{gpt-3.5-turbo}, assigner=\texttt{flan-t5}}
    \label{tab:app-descriptions-agnews}
\end{table*}
\begin{table*}[]
    \centering
\begin{tabular}{lll}
\toprule
                                     reference explanation &                                      output explanation &  F$_1$ \\
\midrule
                ``\textit{has a topic of Company}'' &                ``\textit{has a topic of business}'' &  75 \\
``\textit{has a topic of Educational Institution}'' &               ``\textit{has a topic of education}'' &  91 \\
                 ``\textit{has a topic of Artist}'' &                  ``\textit{has a topic of rivers}'' &   0 \\
                ``\textit{has a topic of Athlete}'' &   ``\textit{has a topic of sports and recreation}'' &  93 \\
          ``\textit{has a topic of Office Holder}'' & ``\textit{has a topic of politics and government}'' &  90 \\
 ``\textit{has a topic of Mean Of Transportation}'' &      ``\textit{has a topic of military equipment}'' &  82 \\
               ``\textit{has a topic of Building}'' &            ``\textit{has a topic of architecture}'' &  82 \\
          ``\textit{has a topic of Natural Place}'' &               ``\textit{has a topic of mountains}'' &  56 \\
                ``\textit{has a topic of Village}'' &      ``\textit{has a topic of villages and towns}'' &  99 \\
                 ``\textit{has a topic of Animal}'' &                   ``\textit{has a topic of lakes}'' &   3 \\
                  ``\textit{has a topic of Plant}'' &                 ``\textit{has a topic of biology}'' &  73 \\
                  ``\textit{has a topic of Album}'' &                   ``\textit{has a topic of music}'' &  75 \\
                   ``\textit{has a topic of Film}'' &                  ``\textit{has a topic of cinema}'' &  89 \\
           ``\textit{has a topic of Written Work}'' &              ``\textit{has a topic of literature}'' &  70 \\
\bottomrule
\end{tabular}
    \caption{DBpedia, clustering based on Topics, proposer=\texttt{gpt-3.5-turbo}, assigner=\texttt{flan-t5}}
    \label{tab:app-descriptions-dbpedia}
\end{table*}
\begin{table*}[]
    \centering
\begin{tabular}{lll}
\toprule
                   reference explanation &                           output explanation &  F$_1$ \\
\midrule
  ``\textit{has a location of iraq}'' &              ``\textit{has a location of Iraq}'' &  63 \\
 ``\textit{has a location of russia}'' &            ``\textit{has a location of Russia}'' &  77 \\
  ``\textit{has a location of japan}'' &             ``\textit{has a location of japan}'' &  86 \\
 ``\textit{has a location of canada}'' &            ``\textit{has a location of canada}'' &  81 \\
``\textit{has a location of britain}'' &           ``\textit{has a location of Britain}'' &  87 \\
 ``\textit{has a location of france}'' &            ``\textit{has a location of France}'' &  82 \\
``\textit{has a location of germany}'' &           ``\textit{has a location of germany}'' &  79 \\
``\textit{has a location of america}'' & ``\textit{has a location of the United States}'' &  48 \\
  ``\textit{has a location of china}'' &             ``\textit{has a location of china}'' &  84 \\
  ``\textit{has a location of italy}'' &             ``\textit{has a location of italy}'' &  94 \\
\bottomrule
\end{tabular}
    \caption{NYT, clustering based on Locations, proposer=\texttt{gpt-3.5-turbo}, assigner=\texttt{flan-t5}}
    \label{tab:app-descriptions-nyt-location}
\end{table*}
\begin{table*}[]
    \centering
\begin{tabular}{lll}
\toprule
                    reference explanation &                                       output explanation &  F$_1$ \\
\midrule
    ``\textit{has a topic of health}'' &                    ``\textit{has a topic of healthcare}'' &  78 \\
    ``\textit{has a topic of estate}'' & ``\textit{has a topic of housing and living situations}'' &  80 \\
  ``\textit{has a topic of politics}'' &               ``\textit{has a topic of war and weapons}'' &  68 \\
   ``\textit{has a topic of science}'' &                ``\textit{has a topic of climate change}'' &  28 \\
    ``\textit{has a topic of sports}'' &        ``\textit{has a topic of sports and competition}'' &  97 \\
  ``\textit{has a topic of business}'' &        ``\textit{has a topic of business and economics}'' &  77 \\
      ``\textit{has a topic of arts}'' &                ``\textit{has a topic of art exhibition}'' &  63 \\
``\textit{has a topic of technology}'' &  ``\textit{has a topic of technology and communication}'' &  60 \\
 ``\textit{has a topic of education}'' &                     ``\textit{has a topic of education}'' &  82 \\
\bottomrule
\end{tabular}
    \caption{NYT, clustering based on Topics, proposer=\texttt{gpt-3.5-turbo}, assigner=\texttt{flan-t5}}
    \label{tab:app-descriptions-nyt-topic}
\end{table*}
\begin{table*}[]
    \centering
\begin{tabular}{lll}
\toprule
                            reference explanation &                             output explanation &  F$_1$ \\
\midrule
``\textit{has a natural language of English}'' & ``\textit{has a natural language of english}'' &  95 \\
 ``\textit{has a natural language of French}'' &  ``\textit{has a natural language of french}'' &  99 \\
``\textit{has a natural language of Deutsch}'' &  ``\textit{has a natural language of german}'' &  96 \\
``\textit{has a natural language of Spanish}'' & ``\textit{has a natural language of spanish}'' & 100 \\
\bottomrule
\end{tabular}
    \caption{\syndataset{}, clustering based on Language, proposer=\texttt{gpt-3.5-turbo}, assigner=\texttt{flan-t5}}
    \label{tab:app-descriptions-syn-language}
\end{table*}
\begin{table*}[]
    \centering
\begin{tabular}{lll}
\toprule
                             reference explanation &                                        output explanation &  F$_1$ \\
\midrule
    ``\textit{has a writing style of twitter}'' & ``\textit{has a writing style of health and wellness advice}'' &  32 \\
``\textit{has a writing style of screen play}'' &          ``\textit{has a writing style of news article}'' &  42 \\
        ``\textit{has a writing style of rap}'' &    ``\textit{has a writing style of instructional text}'' &  22 \\
       ``\textit{has a writing style of poem}'' &  ``\textit{has a writing style of artistic description}'' &  28 \\
\bottomrule
\end{tabular}
    \caption{\syndataset{}, clustering based on Style, proposer=\texttt{gpt-3.5-turbo}, assigner=\texttt{flan-t5}}
    \label{tab:app-descriptions-syn-style}
\end{table*}
\begin{table*}[]
    \centering
\begin{tabular}{lll}
\toprule
                             reference explanation &                                        output explanation &  F$_1$ \\
\midrule
    ``\textit{has a writing style of twitter}'' & ``\textit{has a writing style of instructional or informat...}'' &  45 \\
``\textit{has a writing style of screen play}'' & ``\textit{has a writing style of narrative or storytelling...}'' &  51 \\
        ``\textit{has a writing style of rap}'' & ``\textit{has a writing style of using rhymes and rhythm}'' &  49 \\
       ``\textit{has a writing style of poem}'' & ``\textit{has a writing style of incorporating foreign lan...}'' &  34 \\
\bottomrule
\end{tabular}
    \caption{\syndataset{}, clustering based on Style, proposer=\texttt{gpt-4}, assigner=\texttt{gpt-3.5-turbo}}
    \label{tab:app-descriptions-syn-style-gpt4}
\end{table*}
\begin{table*}[]
    \centering
    \resizebox{\linewidth}{!}{

\begin{tabular}{lll}
\toprule
                                       reference explanation &                                       output explanation &  F$_1$ \\
\midrule
``\textit{has a topic of what sports to do to improve your...}'' & ``\textit{has a topic of sports and physical activity}'' &  98 \\
``\textit{has a topic of a new anime has been announced}'' &          ``\textit{has a topic of anime and animation}'' &  99 \\
``\textit{has a topic of a tech company releases a new gro...}'' &          ``\textit{has a topic of advanced technology}'' &  99 \\
``\textit{has a topic of how to improve your productivity}'' &       ``\textit{has a topic of workplace productivity}'' &  97 \\
\bottomrule
\end{tabular}
}
    \caption{\syndataset{}, clustering based on Topics, proposer=\texttt{gpt-3.5-turbo}, assigner=\texttt{flan-t5}}
    \label{tab:app-descriptions-syn-topic}
\end{table*}

\section{\opendataset{} Datasets} \label{sec:opengoalex}
Most of our problems are adapted from the \textsc{OpenD5} dataset from \citet{zhong2023goal}. 
To save budget, for each corpus we randomly sampled 400 text samples.

\noindent \textbf{human-written-feedback.} human-written feedback for model-generated summaries \cite{scheurer2023training}, with the goal of ``\textit{categorizing model errors}.

\noindent \textbf{abc-headlines}. We collect headlines published by ABC news, an American news company from \citet{abc-headlines}. 
ABC headlines are directly downloaded from Harvard Dataverse. The year is extracted from the publication date field. Samples are constructed from the headline text.
The goal is to cluster based on the topic of the news.
The data is downloadable from \url{https://doi.org/10.7910/DVN/SYBGZL} with license CC0 1.0.

\noindent \textbf{amazon-reviews}. We collect Amazon reviews collected from various product categories from \citet{amazon-reviews}. Amazon reviews are downloaded from a 2018 crawl of the website. 
The goal is to cluster based on what aspects did the customer complained about the product.
The dataset can be downloaded at \url{https://nijianmo.github.io/amazon/index.html}.
We considered three categories: Beauty product, electronics, and office products.

\noindent \textbf{rate-my-prof}. We collect reviews of lecturers from RateMyProfessor.com from \citet{rate-my-prof}. We download a sample of RateMyProfessor.com reviews from an online repo.
The goal is to cluster based on what aspects did the students comment on the teacher. 
This dataset can be downloaded from \url{https://data.mendeley.com/datasets/fvtfjyvw7d/2} under CC BY 4.0 .

\noindent \textbf{debate-arguments} arguments for a position \cite{habernal-gurevych-2016-makes}, with the goal of ``\textit{categorizing the types of arguments}''. We took the subset of arguments for the position ``\textit{why spanking is bad}'', ``\textit{why william farquhar ought not to be honoured as the rightful founder of singapore}'', and ``\textit{"tv is better than books}''. 

\noindent \textbf{clickbait-headlines} We collect headlines across time from the Examiner, a clickbait news site from \citet{clickbait-headlines}. The Examiner headlines are directly downloaded from Kaggle. Samples are constructed from the headline text.
The goal is to cluster based on their topics.
The dataset can be downloaded at \url{https://www.kaggle.com/datasets/therohk/examine-the-examiner}, with license CC0: public domain.

\noindent \textbf{happy-moments}. We collect self-reported happy moments and demographic characteristics from \citet{happy-moments}. The HappyDB dataset is downloaded from the official GitHub repository. Demographic data is cleaned and merged into happy moments. Happy moment descriptions are treated as samples.
The goal is to cluster based on whom did the person spend the  happy moments with.
This dataset can be downloaded at \url{https://github.com/megagonlabs/HappyDB} under unknown license.

\noindent \textbf{yc$-$startups}. We collect descriptions of companies that were part of the Y Combinator startup incubator from \citet{yc-startups}. YCombinator company descriptions are downloaded from a 2022 scrape on GitHub. Only companies with long descriptions are preserved. 
The goal is to cluster based the type of startups.
The dataset can be downloaded from \url{https://www.kaggle.com/datasets/benhamner/y-combinator-companies}. 

\section{Explainability Evaluation Instance} \label{app:expl-instance}

\noindent \textbf{Explanation by \underline{\methodname{}}.} To reduce the workload of the crowdworkers, we only showed them a condensed summary of the explanation. For example, for the explanation:

``\textit{whether this feedback advises adding omitted details; specifically, the feedback points out that certain key details or aspects are missing from the text, which is necessary for a complete understanding. For example, 'The summary should include specific details about why things in their day went wrong.}''

we only showed the workers

``\textit{whether this feedback advises adding omitted details}''.

This reduces our average explanation length to be around 5 words.

\noindent \textbf{Keyword-based Explanations for LDA and Instructor.}
As \methodname{}-generated explanations have around 5 terms on average, for each of the LDA and Instructor generated cluster, we choose 5 terms to represent and explain the cluster.

Instructor is only able to produce clusters of text based on representations. We couple it with a representative term mining method that is commonly used in text mining and taxonomy construction~\cite{zhang2018taxogen,shang2020nettaxo,mekala2020contextualized}.
The method involves first identifying a vocabulary that the representative terms might fall in. This step is usually done by applying AutoPhrase~\cite{shang2018automated} on the entire text corpus and thresholding the unigrams and multigrams found.
Then, for each cluster, we assign each term in the vocabulary a score based on statistical signals specific to that cluster that correspond to popularity, discriminativeness and informativeness\footnote{While this score has slightly different definitions in different papers, we follow the one from ~\citet{shang2020nettaxo}}. This score can also be seen as a generalized version of tf-idf. Finally, for each cluster, we take the top 5 terms as its explanation.

LDA is able to word-based explanations for topic clusters by itself. However, we found the word clusters LDA generated lack in quality, despite stop word pruning and tf-idf reweighting. We therefore first apply LDA to obtain the topic clusters, and then use the same representative term mining method above to find the top 5 terms.

\noindent\textbf{Implementation Details.}
For Autophrase, we use the official implementation at \url{https://github.com/shangjingbo1226/AutoPhrase}, and do not change the distant supervision or stop words list that was provided. We apply a cutoff threshold of 0.5 for unigram and 0.8 for multigram which we tested to work well on heldout data.

For representative term mining, we re-implement the representativeness score from the ~\citet{shang2020nettaxo} by ourself.

For LDA, we use the implementation in \texttt{sklearn}.

\noindent \textbf{Sanity Check on Keyword Based Implementation.}
To ensure that our implementation of keyword-based explanation is reasonable, we applied LDA and Instructor to cluster topics on the English subset of \syndataset{}, a task that we know that they can achieve better performance.
As expected, Instructor achieves 80\% accuracy while LDA achieves 76\%, which is much higher than the performance on \opendataset{}.

\noindent \textbf{HIT Task.} 
We paid crowdworkers \$0.05 for each binary choice of explanations. 
The authors on average can perform 4 HITs per minute, which translates to around \$12/hour of payment.
We recruited Turkers with > 98\% of HIT acceptance rate in the history. 

\section{Additional Example Taxonomy}

\begin{figure}
    \centering
    \includegraphics[width=\columnwidth]{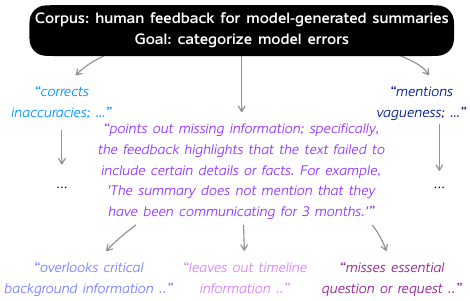}
    \caption{By applying \methodname{} hierarchically to a corpus of human-written comments for model-generated summaries ~\citep{scheurer2023training}, we can automatically induce taxonomies of error categories for a text summarization system. The texts in quotes are copies (sometimes abbreviated) from \methodname{}'s output.}
    \label{fig:demo}
\end{figure}

\begin{figure*}[t!]
    \centering    \includegraphics[width=0.98\linewidth]{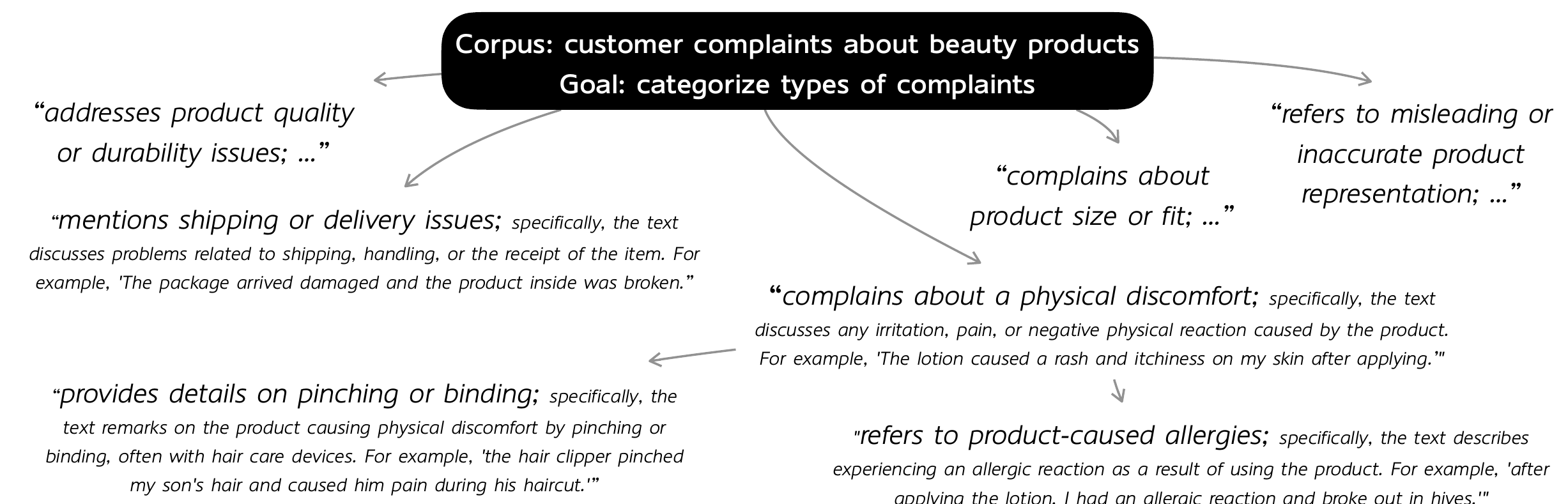}
    \caption{Example taxonomy for complaints about beauty products from Amazon \citep{he2016ups}. }
    \label{fig:product}
\end{figure*}

We provide additional example taxonomy over model errors and customer reviews in Figure \ref{fig:demo} and \ref{fig:product}. 

\section{Implementation Details}\label{app:implementation}

In terms of software libraries, we used \texttt{pulp}  ~\citep{Mitchell2011PuLPA} to implement ILP for \underline{\methodname{}};
we used \texttt{transformers} ~\citep{wolf2020transformers} to run \texttt{flan-t5-xl}, \texttt{e5-large}, and \texttt{instructor-xl};
we used \texttt{sklearn} ~\citep{scikit-learn} to run LDA and K-means on text embeddings.

\section{Sensitivity Study}\label{app:sensitivity}
\begin{table*}[t]
    \small
    \centering
    \begin{tabular}{l|ccccccc}
    \toprule
    \textbf{} Macro F$_1$ (\%) & (AG) & (DB) & (NYT) & (SYN) & (NYT) & (SYN) & (SYN) \\
    Goal & \multicolumn{4}{c}{Topic} & Location & Language & Style \\
    \hline
    \underline{Instructor}    & 84& 82 &69 & 77&54 &25 &25\\
    \underline{\methodname}          & 87& 71 &70 &98 &76 &97 &31 \\
    \underline{\methodname} (new proposer prompt)           & 86&72&68&98&75&98&28 \\
    \underline{\methodname} (new assigner prompt)         &87&63&67&98&82&98&29 \\

    \bottomrule
    \end{tabular}
    \caption{We paraphrase the prompt used for the proposer and assigner and assess their performance. 
    }
    \label{tab:different_prompt}
\end{table*}
\begin{table}[t]
    \small
    \centering
    \begin{tabular}{l|ccc}
    \toprule
    Delta Macro F$_1$ (\%)  & \multicolumn{3}{c}{sample seed} \\
     & seed = 0 & seed = 1 & seed = 2 \\
    \hline
    \multicolumn{4}{c}{\textit{Imbalance}} \\
    \hline
    \underline{Instructor}    & -11 & -14 & -10 \\
    \underline{\methodname}   & -10 & -8 & -12 \\
    \hline
    \multicolumn{4}{c}{\textit{Noise}} \\
    \hline
    \underline{Instructor}    & -1 & -1 & -2 \\
    \underline{\methodname}   & -2 & -3 & -3 \\
    \bottomrule
    \end{tabular}
    \caption{We create three imbalanced and noisy versions of the DBpedia dataset where the difference is at the random seed during data creation. We calculate the difference between the performance on the clean dataset of Instructor and \methodname . 
    }
    \label{tab:imbalance}
\end{table}

\noindent \textbf{Prompt Sensitivity Study.} 
We conduct a study to understand how sensitive our method is to our crafted prompts. We perturb our proposer and assigner prompt template by paraphrasing it extensively; the perturbed prompts are in Appendix~\ref{app:prompt-templates}. The results are shown in Table~\ref{tab:different_prompt}. The performance does not change much when the prompt changes. Most importantly, the two claims, (1) our method is on par with prior clustering methods on topic clustering, and (2) our method is much better than prior methods when the goal is non-topic, still holds robustly.

\noindent \textbf{Dataset Sensitivity Study.}
Real world data is usually not clean and perfectly balanced. To understand how the data balance and data noise affects our method, we conduct the following two studies on the (DB)pedia dataset.
For the imbalanced scenario, we randomly sampled 7 classes and removed half of their data points. Then among the 7 classes, we further random sampled 3 classes and again removed half of their remaining data points. In Table~\ref{tab:imbalance}, we report the delta changes of Instructor and our method PAS, and observe that the change is similar when the dataset becomes imbalanced. 

For the noisy experiment, we randomly sampled 4 classes and removed $\frac{7}{8}$ of their data points. Then, we consider the clustering problem on the 10 remaining classes, but with the additional data of the 4 removed classes as extra noise. The evaluation is only done on the data points of the 10 remaining classes, but the model needs to be robust to noisy data points. From Table~\ref{tab:imbalance}, we observe that both Instructor and our method have a similar small drop in performance.

Both results indicate that our method is not especially vulnerable to imbalance or noise in the dataset.
\label{sec:appendix}

\end{document}